\documentclass[sigconf, nonacm]{acmart}

\usepackage{xcolor}
\usepackage{multirow}
\usepackage{enumitem}
\newcommand{\RR}{\mathbb{R}}
\newcommand{\ZZ}{\mathbb{Z}}

\usepackage{caption}
\usepackage{subcaption}

\usepackage{xspace}
\newcommand{\proj}{SUREL+\xspace}

\usepackage[ruled,linesnumbered]{algorithm2e}

\makeatletter
\newcount\bracketnum
\newcommand\makecolorlist[1]{%
    \bracketnum0\relax
    \makecolorlist@#1,.%
    \bracketnum0\relax
}
\def\makecolorlist@#1,{%
    \advance\bracketnum1\relax
    \expandafter\def\csname bracketcolor\the\bracketnum\endcsname{\color{#1}}%
    \@ifnextchar.{\@gobble}{\makecolorlist@}%
}
\let\oldleft\left
\let\oldright\right
\def\left#1{%
    \global\advance\bracketnum1\relax 
    \colorlet{temp}{.}%
    \csname bracketcolor\the\bracketnum\endcsname
    \oldleft#1%
    \color{temp}%
}
\def\right#1{%
    \colorlet{temp}{.}%
    \csname bracketcolor\the\bracketnum\endcsname
    \oldright#1%
    \global\advance\bracketnum-1\relax
    \color{temp}%
}
\makeatother

\makecolorlist{black,blue,orange,red}

\theoremstyle{definition}
\newtheorem{definition}{Definition}[section]

\newcommand\vldbdoi{10.14778/3611479.3611499}
\newcommand\vldbpages{2939-2948}
\newcommand\vldbvolume{16}
\newcommand\vldbissue{11}
\newcommand\vldbyear{2023}
\newcommand\vldbauthors{Haoteng Yin, Muhan Zhang, Jianguo Wang, Pan Li}
\newcommand\vldbtitle{\shorttitle} 
\newcommand\vldbavailabilityurl{https://github.com/Graph-COM/SUREL_Plus}
\newcommand\vldbpagestyle{plain} 

\begin{document}
\title{SUREL+: Moving from Walks to Sets for Scalable Subgraph-based Graph Representation Learning}

\author{Haoteng Yin$^{\dagger}$, Muhan Zhang$^{\ddagger}$, Jianguo Wang$^\dagger$, Pan Li$^{\dagger\S}$}
\affiliation{
{{$^\dagger$}{Department of Computer Science, Purdue University}}~~~~~$^\ddagger$Institute for Artificial Intelligence, Peking University \\{{$^{\S}$}{School of Electrical and Computer Engineering, Georgia Institute of Technology}}}
\affiliation{
$^\dagger$\{yinht, csjgwang\}@purdue.edu $^\ddagger$muhan@pku.edu.cn $^{\S}$panli@gatech.edu
}

\begin{abstract}
Subgraph-based graph representation learning (SGRL) has recently emerged as a powerful tool in many prediction tasks on graphs due to its advantages in model expressiveness and generalization ability. Most previous SGRL models face computational challenges associated with the high cost of subgraph extraction for each training or test query. Recently, SUREL was proposed to accelerate SGRL, which samples random walks offline and joins these walks online as a proxy of subgraph for representation learning. Thanks to the reusability of sampled walks across different queries, SUREL achieves state-of-the-art performance in terms of scalability and prediction accuracy. However, SUREL still suffers from high computational overhead caused by node duplication in sampled walks. In this work, we propose a novel framework \proj that upgrades SUREL by using node sets instead of walks to represent subgraphs. This set-based representation eliminates repeated nodes by definition but can also be irregular in size. To address this issue, we design a customized sparse data structure to efficiently store and access node sets and provide a specialized operator to join them in parallel batches. \proj is modularized to support multiple types of set samplers, structural features, and neural encoders to complement the structural information loss after the reduction from walks to sets. Extensive experiments have been performed to validate \proj in the prediction tasks of links, relation types, and higher-order patterns. \proj achieves 3-11$\times$ speedups of SUREL while maintaining comparable or even better prediction performance; compared to other SGRL baselines, \proj achieves $\sim$20$\times$ speedups and significantly improves the prediction accuracy.
\end{abstract}

\maketitle

\pagestyle{\vldbpagestyle}
\begingroup\small\noindent\raggedright\textbf{PVLDB Reference Format:}\\
\vldbauthors. \vldbtitle. PVLDB, \vldbvolume(\vldbissue): \vldbpages, \vldbyear.\\
\href{https://doi.org/\vldbdoi}{doi:\vldbdoi}
\endgroup
\begingroup
\renewcommand\thefootnote{}\footnote{\noindent
This work is licensed under the Creative Commons BY-NC-ND 4.0 International License. Visit \url{https://creativecommons.org/licenses/by-nc-nd/4.0/} to view a copy of this license. For any use beyond those covered by this license, obtain permission by emailing \href{mailto:info@vldb.org}{info@vldb.org}. Copyright is held by the owner/author(s). Publication rights licensed to the VLDB Endowment. \\
\raggedright Proceedings of the VLDB Endowment, Vol. \vldbvolume, No. \vldbissue\ %
ISSN 2150-8097. \\
\href{https://doi.org/\vldbdoi}{doi:\vldbdoi} \\
}\addtocounter{footnote}{-1}\endgroup

\ifdefempty{\vldbavailabilityurl}{}{
\vspace{.3cm}
\begingroup\small\noindent\raggedright\textbf{PVLDB Artifact Availability:}\\
The source code, data, and/or other artifacts have been made available at \url{\vldbavailabilityurl}.
\endgroup
}

\section{Introduction \label{sec:0}}
\begin{figure*}
    \centering
    \begin{minipage}{0.26\textwidth}
        \centering
        \includegraphics[width=0.78\textwidth]{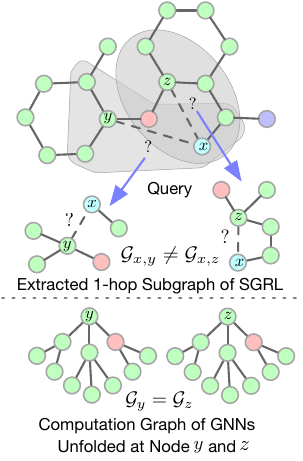}
        \vspace{-5mm}
        \caption{\small{GNNs cannot correctly predict whether $x$ is more likely linked with $y$ or $z$: $y$ and $z$ have the same representation without distinct features. The representations based on one-hop neighbors (query-induced subgraph) are more expressive in distinguishing node pairs $(x,y)$ and $(x,z)$.}\label{fig:sgrl}}
    \end{minipage}\hfill
    \begin{minipage}{0.72\textwidth}
        \centering
        \includegraphics[width=\textwidth]{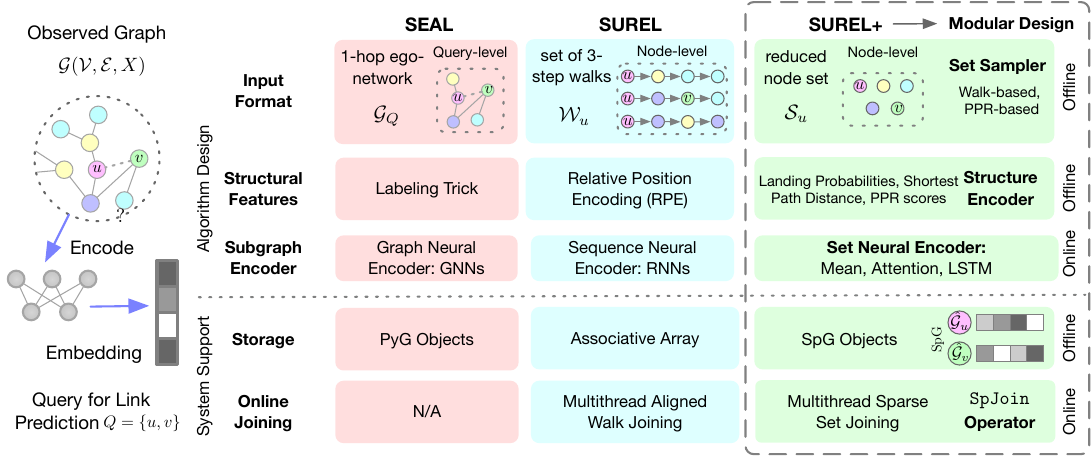}
        \vspace{-8mm}
        \caption{\small{Overview of \proj: a side-by-side comparison with previous SGRL models. For a query, SGRL first prepares the input subgraph attached with structural features, which are fed into a neural encoder to obtain embeddings for prediction. \proj samples node sets as input, while SEAL~\cite{zhang2018link,zhang2021labeling} extracts query-induced subgraphs and SUREL~\cite{yin2022algorithm} runs random walks. To compensate for the structure loss by reducing subgraphs to node sets, \proj provides various types of set samplers, structure encoders, and set neural encoders. To better serve set-based representations, \proj designs a customized sparse data structure \texttt{SpG} to store sampled node sets with fast access, and supports their online joining in parallel via a sparse join operator \texttt{SpJoin}.}\label{fig:main}}
    \end{minipage}
    \vspace{-3mm}
\end{figure*}

Graphs are widely used to model interactions in natural sciences and relationships in social life~\cite{koller2007introduction,jumper2021highly}. Graph-structured data in the real world are highly irregular and often large-scale. To solve inference tasks on graphs, graph representation learning (GRL) that studies quantitative representations of graph-structured data has attracted much attention~\cite{hamilton2017representation,hamilton2020graph,wu2022graph}. Recently, subgraph-based GRL (SGRL) has become an important research direction for researchers studying GRL algorithms and systems, as it achieves far better prediction performance than other approaches on many GRL tasks, especially those involving a set of nodes. Given a set of queried nodes, SGRL models such as SEAL~\cite{zhang2018link,zhang2021labeling}, GraIL~\cite{teru2020inductive}, and SubGNN~\cite{alsentzer2020subgraph} first extract a subgraph around the queried node set (called query-induced subgraph), and then use neural networks to encode extracted subgraphs for prediction. Extensive work shows that SGRL models are more robust~\cite{zeng2021decoupling} and more expressive~\cite{bouritsas2022improving,frasca2022understanding}; while canonical graph neural networks (GNNs) including GCN~\cite{kipf2016semi} and GraphSAGE~\cite{hamilton2017inductive} usually fail to make accurate predictions, due to their limited expressive power~\cite{zhang2021labeling,garg2020generalization,chen2020can}, incapability of capturing intra-node distance information~\cite{srinivasan2019equivalence,li2020distance}, and improper entanglement between receptive field size and model depth~\cite{huang2020graph,zeng2021decoupling,yin2022algorithm}. An example in Fig. \ref{fig:sgrl} illustrates how SGRL works for link prediction and demonstrates its advantages over GNNs. Here, canonical GNNs generate and aggregate node-wise representations to predict links, which would map structurally symmetric nodes without distinct features into the same representation and thus lead to the ambiguity issue \cite{xu2019powerful,zhang2021labeling}. So far, the advantages of SGRL methods have been verified in many applications, such as link and relation prediction~\cite{zhang2018link,zhang2021labeling,teru2020inductive}, higher-order pattern prediction~\cite{meng2018subgraph,liu2021neural}, temporal network modeling~\cite{wang2021inductive}, recommender systems~\cite{zhang2020inductive}, anomaly detection~\cite{alsentzer2020subgraph,cai2021structural}, graph meta-learning~\cite{huang2020graph}, subgraph matching~\cite{liu2020neural,lou2020neural}, and molecular/protein study in life sciences~\cite{wang2021glass}. 

\begin{figure}[t]
\centering
\includegraphics[width=0.47\textwidth]{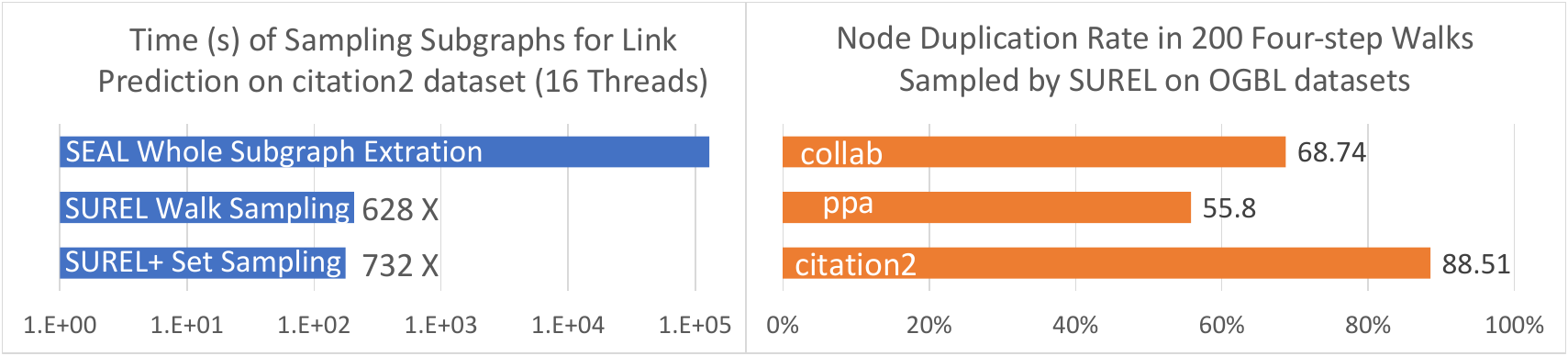}
\vspace{-4mm}
\caption{\small{(a) Subgraph extraction in SEAL~\cite{zhang2018link,zhang2021labeling} has much higher complexity than samplers of other simplified forms. The size of query-induced subgraphs is irregular, growing exponentially w.r.t. the number of extracted hops~\cite{chen2018fastgcn,zeng2021decoupling}. (b) Breaking subgraphs into reusable walks faces a serious node duplication issue.} \label{fig:subg_sampler}}
\vspace{-4mm}
\end{figure}

Albeit with multiple benefits of its algorithm, SGRL methods currently face two major computational challenges: (1) \textbf{Query Dependency}. A subgraph must be extracted for each queried node set, which is not reusable across different queries, and cannot be preprocessed if the query is unknown; (2) \textbf{Irregularity}. The extracted subgraphs are irregularly sized, resulting in poor batch processing and load-balancing performance. As Fig. \ref{fig:subg_sampler} (a) shows, subgraph extraction in SEAL~\cite{zhang2018link,zhang2021labeling} is prohibitively slow for practical deployment. This inspired recent work on dedicated hardware acceleration for extracting subgraphs~\cite{dgl67,pyg22}. However, how to fundamentally improve the scalability and efficiency of SGRL methods remains largely unexplored. 

SUREL \cite{yin2022algorithm} is the state-of-the-art (SOTA) framework that applies algorithm and system co-design to implement SGRL. 
It decouples the input from specific queries by sampling node-level random walks and uses the joint walks of queried nodes as a proxy of subgraphs. Specifically, SUREL treats each node in the graph as a seed and runs multiple random walks from the seed offline. Given a queried node set, SUREL online joins and encodes the sampled walks of all queried nodes for prediction. The join operation builds the connection between sampled walks of queried nodes so that their joins can function as the query-induced subgraph. To compensate for the structure loss by representing subgraphs in walks, a structural feature termed relative position encoding (RPE) is adopted to record the relative distance between each distinct node in sampled walks and the seed. The RPE is pre-computed offline and attached to sampled walks before being fed into neural networks (NNs) to make predictions. Sampled walks from one seed can be reused for multiple queries whenever that seed node is involved. Through this walk-sharing mechanism, SUREL significantly improves the efficiency of SGRL. The regularity of walks also enables highly parallel walk sampling and online joining with a dedicated system design. However, SUREL still faces several inherent drawbacks of the walk-based representation, namely high node redundancy in sampled walks (over 55\% is duplicated, see Fig. \ref{fig:subg_sampler} (b)). This further raises the following issues: (1) \textbf{extra space} for hosting walks and positional encoding in memory, (2) \textbf{extra time} for operations on duplicated nodes in subsequent routines of walk joining and NN-based encoding, and (3) \textbf{high workload} of data transfer between CPU and GPU.

In this work, we upgrade SUREL and develop a novel framework \proj that again benefits from algorithm and system co-design. The core concept of \proj is simple: instead of using walks to represent subgraphs, we now employ node sets, thereby obviating node duplication. However, this new idea based on node sets also introduces several algorithm and system design difficulties. From an algorithmic perspective, as explored in this study, the transition from induced subgraphs to walks and then to node sets results in a considerable loss of structural information. Therefore, the first priority is to develop a method that can compensate for such loss while maintaining performance. On the system side, SUREL~\cite{yin2022algorithm} utilizes walks, which can be easily stored and processed in an aligned format by controlling sampling parameters. In contrast, node sets are irregular in size, creating difficulties for efficient storage and fast access. To sum up, how to coordinate the designs of both sides constitutes the primary challenge of this work. 

\proj tackles the above challenges through its entire pipeline. To avoid node duplication in walk sampling, \proj only keeps unique nodes from neighborhood sampling of seed nodes during preprocessing. To compensate for structural information loss, \proj incorporates various types of set samplers and structure encoders to preserve local graph structures: \emph{set samplers} employ different graph metrics to measure node importance and determine sampling rules; \emph{structure encoders} support landing probabilities of random walks~\cite{li2019optimizing}, shortest path distances, and personalized PageRank scores~\cite{jeh2003scaling}, covering most of the structural features used by previous SGRL models~\cite{zhang2018link,zhang2021labeling,li2020distance,teru2020inductive,yin2022algorithm}. Furthermore, \proj designs a customized sparse data structure, namely \texttt{SpG}, which can efficiently store sampled node sets and achieve fast access. A sparse operator \texttt{SpJoin} is developed accordingly to perform join operations on the sampled node sets and associated structural features for serving queries online. To capture diverse levels of interactions between node and structural features, \proj introduces multiple \emph{set neural encoders}, such as multi-linear perception with mean pooling, set attention~\cite{velivckovic2017graph} and LSTM~\cite{hamilton2017inductive} that ensure sufficient expressive power and consistent performance across various types of SGRL tasks. 

Overall, our contributions can be summarized as follows: 
\begin{itemize}[leftmargin=*]
    \item Algorithm: \proj is a novel SGRL framework (\emph{open source}), which utilizes reusable node sets associated with various structural features to represent query-induced subgraphs via online joining. Compared with the SOTA baselines, the proposed set-based subgraph representation greatly reduces memory and computation costs without degrading prediction performance.
    \item System: \proj designs a customized sparse data structure \texttt{SpG} and a sparse join operator \texttt{SpJoin} to support efficient storage and fast access of node sets, which achieves much lower latency and higher throughput than previous SGRL methods.
    \item We conduct extensive experiments on 9 real-world graphs, with millions/billions of nodes/edges, and demonstrate the advantages of \proj in link/relation-type/motif prediction tasks. \proj is 3-11$\times$ faster than the current SOTA SGRL method SUREL while maintaining comparable or even better prediction accuracy. \proj also achieves $\sim$20$\times$ speedups with substantial accuracy improvements over other SGRL baselines.
\end{itemize}
\section{Preliminaries}
\subsection{Notations and Relevant Definitions in SGRL}
Let $\mathcal{G}(\mathcal{V},\mathcal{E},X)$ be an attributed graph with node set $\mathcal{V}=\{1,2,...,n\}$ and edge set $\mathcal{E}$, where $X \in \RR^{n\times d}$ denotes node attributes with $d$-dimension. A query $Q \subset \mathcal{V}$ is a node set of interest for a certain type of task. We denote the subgraph induced by query $Q$ as $\mathcal{G}_Q$ and the node-induced subgraph as $\mathcal{G}_u$, where induced subgraphs are typically within a small number of hops.

\begin{definition}[Subgraph-based Graph Representation Learning (\textbf{SGRL})\label{def:SGRL}] Given a query $Q$ of node set over graph $\mathcal{G}$, SGRL aims to learn a representation of the query-induced subgraph $\mathcal{G}_Q$ to make prediction $f(\mathcal{G}_Q)$. $f(\cdot)$ is usually a neural network. SGRL tasks come with some labeled queries $\{(Q_i,y_i)\}_{i=1}^L$ for supervision (positive samples) and other unlabeled queries $\{Q_i\}_{i=L+1}^{L+N}$ for inference.
\end{definition}

\textbf{Examples of SGRL Tasks} \emph{Link prediction} seeks to estimate the likelihood of a link between two nodes in a given graph, where a query $Q$ corresponds to a node pair. It can be further generalized to predict links with types over heterogeneous graphs~\cite{teru2020inductive} or to predict blood vessels~\cite{paetzold2021whole} and chemical bonds~\cite{jumper2021highly} in domain-specific graphs. Tasks beyond pairwise relations are named \emph{higher-order pattern prediction}, where a query $Q$ consists of three or more nodes. In this work, we consider that given partially observed pairwise relations among queried nodes in $Q$, whether these queried nodes will establish certain full higher-order relation of interest~\cite{srinivasan2021learning,liu2021neural}.

\textbf{Review of SGRL Methods} The current SGRL pipeline mainly has three parts, as shown in the \emph{Algorithm Design} section of Fig. \ref{fig:main}: subgraph preparation, structural feature construction, and neural encoder to obtain the readout of subgraphs. Classical SGRL models often group query-dependent parts together (e.g., SEAL~\cite{zhang2018link,zhang2021labeling} couples subgraph extraction with labeling trick~\cite{zhang2021labeling}), and then apply GNNs on extracted and labeled subgraphs for prediction. However, such coupling is expensive and makes the computed intermediate results (e.g. labeled subgraphs) not reusable across queries, which motivates recent SGRL methods to decouple them. SUREL \cite{yin2022algorithm} substitutes explicit subgraph extraction with online joining of multiple pre-sampled walks attached with positional encoding defined on walk landing as structural features, both of which are node-level and thus can be reused to serve multiple queries. Lastly, it applies neural networks to encode joint walks and aggregate their embeddings for prediction.

\vspace{-0.5mm}
\subsection{Related Works}
\textbf{Scalable SGRL Design.} Recent works on SGRL models have primarily focused on efficient subgraph extraction. Various techniques have been proposed, including PPR-based~\cite{bojchevski2020scaling,zeng2021decoupling} and random walk-based~\cite{yin2022algorithm} subgraph samplers, node neighborhood sampling through CUDA kernel (DGL, ~\cite{dgl67}) and tensor operations (PyG,~\cite{pyg22}). Some frameworks have customized data structures to better support subgraph operations and gain higher throughput, such as associative arrays in SUREL~\cite{yin2022algorithm}, temporal-CSR in TGL~\cite{zhou2022tgl}, and GPU-orientated dictionary in NAT~\cite{luo2022neighborhood}. To achieve scalable modeling design, GDGNN~\cite{kong2022geodesic} utilizes node representations along the geodesic path between queried nodes for prediction, partially decoupling structural feature construction from subgraph extraction. BUDDY~\cite{chamberlain2022graph} employs subgraph sketches to avoid explicitly constructing subgraphs for link prediction. However, these works either focus on specific aspects of computational issues in SGRL, namely bottlenecks of extraction, storage, and feature construction, or are limited to one type of SGRL task. In contrast, \proj provides a comprehensive co-design approach in scalable sampling, efficient storage, and expressive modeling, offering a general and scalable framework for various SGRL tasks.
\vspace{-0.5mm}
\section{The framework of \proj}
This section introduces \proj, whose key concept is to sample node sets and encode structural features offline and then join them online as a proxy of query-induced subgraph for representation learning. This approach only keeps distinct nodes in the sampled set for reuse in different queries, effectively addressing memory and computation concerns of node duplication in the walk-based representation adopted by SUREL~\cite{yin2022algorithm}. \proj features a modular design that supports various set samplers, structure encoders, and set neural encoders to provide a trade-off between complexity and expressiveness after reducing subgraphs to node sets. Furthermore, \proj introduces a customized sparse data structure \texttt{SpG} and an arithmetic operator \texttt{SpJoin} to store node sets and perform their online joins efficiently. Fig. \ref{fig:main} summarizes and compares \proj and current SGRL models. The following subsections describe these modules in detail.

\vspace{-1mm}
\subsection{Set Samplers and Structure Encoders}
\label{sec:set-sampler}
\proj uses set samplers to sample a set of nodes from each node's neighborhood and calls structure encoders to construct the corresponding structural features. Both operations are executed offline: the former is primarily for computational benefits, while the latter is to offset the structure loss of node sets reducing from subgraphs (adopted by SEAL~\cite{zhang2018link,zhang2021labeling}) or walks (adopted by SUREL~\cite{yin2022algorithm}). Conceptually, \proj represents the node-induced subgraph $\mathcal{G}_u$ via a combination of (1) a node set $\mathcal{S}_u$ comprising unique nodes sampled from the neighborhood of node $u$ and (2) the associated structural features $\mathcal{Z}_u$ reflects the position of sampled nodes in $\mathcal{G}_u$.

\textbf{Set Samplers} Two types of set samplers are adopted. The first type, named \emph{Walk-based Sampler}, is to sample short-step random walks and then reduce sampled walks into a set of unique nodes. The second type, named \emph{Metric-based Sampler}, is based on graph metrics that measure the proximity between neighboring nodes and the seed, such as personalized PageRank (PPR) scores~\cite{jeh2003scaling} or short path distances. Specifically, the walk-based sampler runs $M$-many $m$-step random walks, starting from each seed $u$ in parallel on the graph $\mathcal{G}$, and then only puts distinct nodes of sampled walks into the set $\mathcal{S}_u$. The metric-based sampler, taking PPR-based \cite{bojchevski2020scaling} as an example, first runs the push-flow algorithm \cite{andersen2006local} to obtain an approximation of the PPR vector for each seed $u$, and then selects the top-$K$ nodes with the highest PPR scores into the set $\mathcal{S}_u$. Mathematically, PPR scores are convergent landing probabilities of seeded random walks that reach infinite steps. Therefore, these two samplers complement each other by leveraging either more local or global structures of the graph. We use hyper-parameters $M$, $m$ to control random walks, and $K$ to control metric-based samplers, which are all set as some constants in practice. The complexity of the above offline sampling procedures is $O(|\mathcal{V}|)$.

\textbf{Structure Encoders} 
The structure encoder is to construct structural features $\mathcal{Z} _{u,x} \in \RR^k$ for each node $x$ in the sampled node set $\mathcal{S}_u$. These features prove to be crucial for inference tasks on graphs involving multiple nodes~\cite{zhang2021labeling} and can be conceptually understood as anchoring sampled node $x$ in the seed $u$'s neighborhood. One possible choice is landing probabilities of random walk \cite{li2019optimizing,li2020distance,yin2022algorithm}: each element $\mathcal{Z}_{u,x}[i]$ stores the counts of node $x$ landed at step $i$ of all walks rooted at the seed $u$ divided by the number of walks performed by the sampler. Landing probabilities (LPs) can be computed along with walk sampling. Another option is the shortest path distance (SPD) between $x$ and $u$ \cite{zhang2018link,li2020distance,zhang2021labeling}, which records their relative distance in terms of reachability. PPR scores~\cite{jeh2003scaling} is also a popular structural feature and can be naturally obtained by running a PPR-based sampler. Later, we denote the collection of structural features for all nodes in $\mathcal{S}_u$ as $\mathcal{Z}_u = \{\mathcal{Z}_{u,x} | x \in \mathcal{S}_u\}$.

\begin{figure}[t]
\centering
\includegraphics[width=0.48\textwidth]{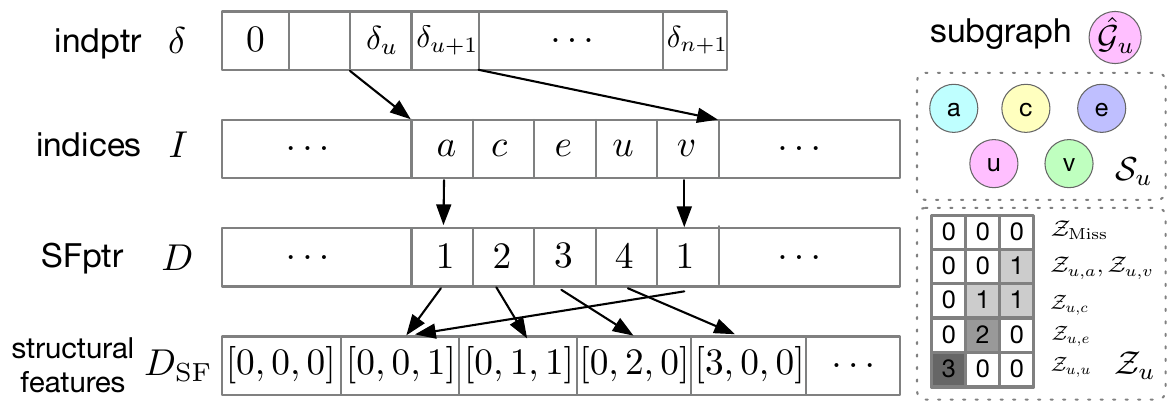}
\vspace{-6mm}
\caption{\small{Node set $\mathcal{S}_u$ and its associated structural features $\mathcal{Z}_u$ stored in \texttt{SpG}. It contains three arrays to store the size of node sets, sampled node indices, and associated structural features (with optional two-level indexing). Here, $D_{\text{SF}}$ shows the landing counts of nodes at different steps in sampled walks as an example, which can be normalized later to landing probabilities as structural features.}}
\label{fig:subg_csr}
\vspace{-5mm}
\end{figure}

\subsection{Set-based Storage - \texttt{SpG}\label{sec:storage}}
Set-based subgraph representation has advantages in terms of flexibility and compactness. However, the uneven sizes of sampled node sets pose great challenges to their storage and fast access. Note that, these node sets need to be frequently visited in subsequent online phases. To overcome these obstacles, \proj designs a customized compressed sparse row (CSR) format called \texttt{SpG}, which reorganizes the storage of node sets and their structural features in a memory-efficient manner, as depicted in Fig. \ref{fig:subg_csr}. Specifically, the node set $\mathcal{S}_u$ and its structural features $\mathcal{Z}_u$ are stored as a row of \texttt{SpG}, denoted as $\texttt{SpG}[u,:]$. Multiple node sets and their associated structural features are consolidated into three contiguous arrays:

\begin{itemize}[leftmargin=*]
    \item \texttt{indptr} $\delta \in \ZZ^{n+1}$, an integer array tracks the starting index of each stored node set (row). It records the cumulative sum of the sizes of all node sets $\mathcal{S}_u,~\forall u \in \mathcal{V}$, e.g., $\delta_{u+1} = \delta_{u} + |\mathcal{S}_u|$, where $|\mathcal{S}_u|$ represents the size of the set $\mathcal{S}_u$. The total number of sampled nodes stored in \texttt{SpG} is $\delta_{n+1}$;
    \item \texttt{indices} $I \in \ZZ^{\delta_{n+1}}$, a coalesce array of all node sets $\mathcal{S}_u,~\forall u\in \mathcal{V}$. The segment $I[\delta_u:\delta_{u+1}]$ corresponds to node indices of the set $\mathcal{S}_u$ in sorted order. This ordering is particularly useful for speeding up the join operation discussed in Sec.~\ref{sec:join}. 
    \item \texttt{SFptr} $D \in \RR^{\delta_{n+1}}$, an array contains the values of the structural features $\mathcal{Z}_u$, or the indices of encoding stored in the array $D_{\text{SF}}$. $D_{\text{SF}}$ is introduced to eliminate duplicate encoding of structural features, which typically reside in GPU memory. This two-level indexing can further reduce memory needs when LPs/SPDs are used, as they are likely to have many repeated values, but it is not necessary when using PPR scores since their values tend to be distinct.
\end{itemize}

\begin{table}[tp]
    \centering
    \caption{\small{Complexity comparison of GRL models. Suppose using $O(|\mathcal{E}|)$-many queries, SGRLs use partial edges ($q\ll |\mathcal{E}|$) for training. $S$ and $K$ denote the average size of extracted subgraphs and sampled node sets, respectively. $L$ is the number of layers. $d$ and $k$ are respective dimensions of node and structural features. Assume $d$ is fixed for all layers. Both SUREL and \proj use the walk-based sampler for $M$-many $m$-step walks. $c$ is the number of distinct $k$-dim structural features. $\delta_{n+1}$ is the size sum of all node sets.}\label{tab:store}}
    \vspace{-2mm}
    \label{tab:prep}
    \resizebox{0.48\textwidth}{!}{
    \begin{tabular}{c|c|c|c|c}
        \toprule 
        \textbf{Methods} & GNN~\cite{kipf2016semi} & SEAL~\cite{zhang2018link,zhang2021labeling} & SUREL~\cite{yin2022algorithm} & \proj \\
        \hline 
        Structure & $O(|\mathcal{V}|+|\mathcal{E}|)$ & $O(S|\mathcal{E}|)$ & $O(mM|\mathcal{V}|)$ & $O(\delta_{n+1})$\\
        Feature & $O(d|\mathcal{V}|)$ & $O(kS|\mathcal{E}|)$ & $O(\delta_{n+1}*k)$ & $O(\delta_{n+1}+c*k)$\\\midrule
        Time & {\bf $O(|\mathcal{E}|Ld + |\mathcal{E}|Ld^2 )$}& $O(qS^Ld^2)$ & $O(qmMd^2)$ & $O(qKd^2)$\\
    \bottomrule
    \end{tabular}}
    \vspace{-2mm}
\end{table}

Regarding the cost of \texttt{SpG}, \texttt{indptr} array is of size $|\mathcal{V}| + 1$, and the size of both \texttt{indices} and \texttt{SFptr} arrays is $\delta_{n+1}$. The compressed encoding array $D_{\text{SF}}$ has a size of $c*k$, where $c$ is the number of distinct structural features and $k$ denotes feature dimension. The overall complexity of \texttt{SpG} is $O(|\mathcal{V}|+\delta_{n+1}+c*k)$.

\textbf{Comparison with Other Methods} Table \ref{tab:store} summarizes the space and time complexity of GRL methods. By adopting the walk-based sampler (sampling $M$-many $m$-step walks), $\delta_{n+1}$ amounts to around one-fifth space of $O(mM|\mathcal{V}|)$ used by SUREL. The metric-based sampler (sampling top-$K$ PPR scores) results in $\delta_{n+1}=K|\mathcal{V}|$ and $K <mM$ in general. Both values are substantially lower than $ O(S|\mathcal{E}|)$ used by SEAL, where $S$ is the average size of extracted subgraphs. \proj further reduces the memory footprint, when the two-level indexing is employed for hosting structural features and only distinct values are stored in $D_{\text{SF}}$. In practice, $c$ typically remains independent of $|\mathcal{V}|$. \texttt{SpG} enables \proj to handle SGRL tasks more efficiently on large-scale graph data.

\subsection{Joining Node Sets via Sparse Operations} \label{sec:join}
\begin{figure}[t]
\centering
\includegraphics[width=0.48\textwidth]{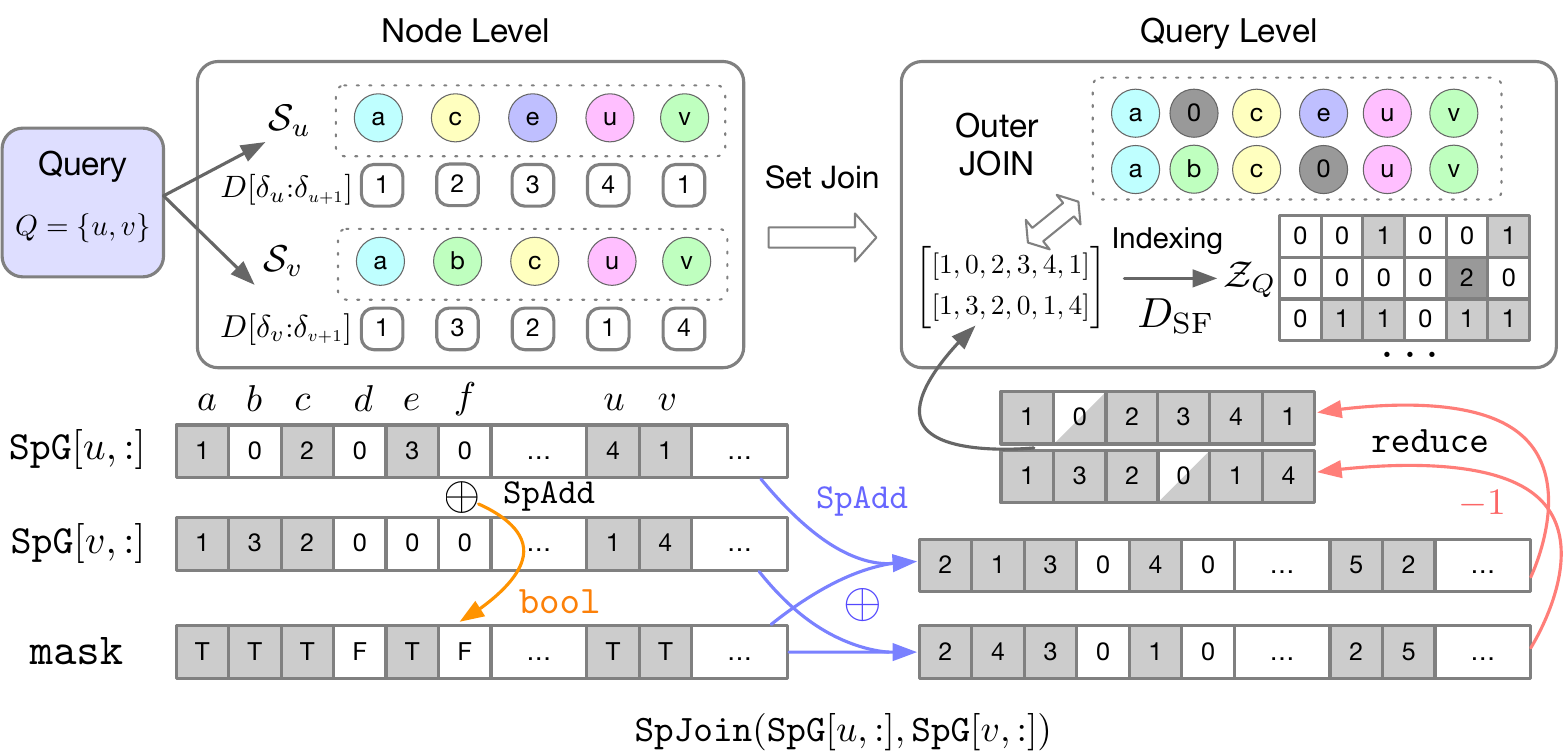}
\vspace{-6mm}
\caption{\small{An illustration of joining structural features from node-level to query-level (Eq.~\eqref{eq:concat}) via the \texttt{SpJoin} operator on node sets stored in \texttt{SpG}. Note that \texttt{SpG} objects do not physically carry $0$ as above shown in $\texttt{SpG}[u,:]$ and $\texttt{SpG}[v,:]$. Only non-zero elements in grey blocks are stored in \texttt{SpG} and performed arithmetic operations by \texttt{SpJoin}. The half-grey blocks correspond to added missing values.}}
\label{fig:sp_join}
\vspace{-3mm}
\end{figure}

The goal of joining node sets is to connect queried nodes and construct query-level subgraphs from pre-sampled node sets to make predictions. For a given query $Q$, we merge relevant node sets $S_u, \forall u \in Q$ into $\mathcal{S}_Q=\bigcup_{u\in Q} \mathcal{S}_u$ and \emph{join} their node-level structural features $\mathcal{Z}_u$ to the query level. In essence, query-level structural features $\mathcal{Z}_Q$ record the relative position of each node $x\in \mathcal{S}_Q$ with respect to the set of queried nodes $Q$ (equivalently labeling the query-induced subgraph $\mathcal{G}_Q$). Specifically, for a node $x$ in $\mathcal{S}_Q$, the query-level structural feature $\mathcal{Z}_{Q,x}$ is obtained by merging its node-level ones $\mathcal{Z}_{u,x}$ for all queried node $u$ in $Q$ as
\begin{align}\label{eq:concat}
\mathcal{Z}_{Q,x} = ||_{u\in Q}\mathcal{Z}_{u,x}=[\dots \mathcal{Z}_{u,x} \dots] \in \RR^{|Q|\times k}, 
\end{align}
where $||$ denotes concatenation. In cases where $\mathcal{Z}_{u,x}$ does not exist as node $x \notin S_u$, it is set to all zeros. For instance, in Fig. \ref{fig:sp_join}, node $b$ is in $\mathcal{S}_v$ but not in $\mathcal{S}_u$, hence $\mathcal{Z}_{u,b}$ is set to zero. $\mathcal{Z}_Q$ is a collection of $\mathcal{Z}_{Q,x}, \forall x\in \mathcal{S}_Q$. Together, $\mathcal{S}_Q$ and $\mathcal{Z}_{Q}$ function as the query-induced subgraph $\mathcal{G}_{Q}$, which is later fed into the neural encoder to obtain the query-level readout for prediction.

The JOIN operator in databases is used to merge tables and establish connections. Concatenation in Eq.~\eqref{eq:concat} requires matching among different node sets with varying sizes and arbitrary node orders, where an outer JOIN is well-suited for this task. In this case, the JOIN operator returns associated values from target sets based on node indices as the specified common field, regardless of their existence. To obtain $\mathcal{Z}_{Q,x}$, node sets $\{\mathcal{S}_u\}_{u \in Q}$ are treated as tables: if the index of node $x$ matches one of the node indices in $\mathcal{S}_u$ for all $u$ in $Q$, then the associated structural feature $\mathcal{Z}_{u,x}$ is appended; otherwise, the field is filled with zeros. However, iterating over all $\mathcal{S}_u$'s to retrieve $\mathcal{Z}_{u,x}$ for each node $x\in \mathcal{S}_Q$ is highly inefficient, as its complexity can be $O(|Q|*|\mathcal{S}_Q|^2)$ per query. This becomes even more challenging when performing these operations for massive queries with varying sizes of $\mathcal{S}_u$ and $\mathcal{S}_Q$. 

To tackle this issue, we design an efficient arithmetic operator \texttt{SpJoin} to perform joins on sparse data objects of \texttt{SpG} in parallel. This operator reduces the per-query time complexity to $O(|Q|*|\mathcal{S}_Q|)$ by taking advantage of node indices of $\mathcal{S}_u$ stored in \texttt{SpG} are unique and in sorted order. The following demonstrates the use of \texttt{SpJoin} for a query $Q=\{u,v\}$.

\textbf{Sparse Join Operator} The operator \texttt{SpJoin} performs an outer JOIN for query $Q$ on the sampled node sets from seeds $u$ and $v$ stored in \texttt{SpG} as $\texttt{SpG}[u,:]$ and $\texttt{SpG}[v,:]$ through
\begin{align*}
\texttt{SpJoin}&(\texttt{SpG}[u,:], \texttt{SpG}[v,:])=\\
&{\color{blue}\texttt{SpAdd}}\left(\texttt{mask}, \texttt{SpG}[u,:]\right){\color{red}-1}~||~ {\color{blue}\texttt{SpAdd}}\left(\texttt{mask}, \texttt{SpG}[v,:]\right){\color{red}-1},
\end{align*}
where $\texttt{mask}={\color{orange}\texttt{bool}}(\texttt{SpAdd}(\texttt{SpG}[u,:],\texttt{SpG}[v,:]))$. 

As illustrated in Fig.~\ref{fig:sp_join}, \texttt{SpJoin} consists of three steps:
\begin{enumerate}[leftmargin=*]
    \item It utilizes sparse arithmetic operations from SciPy \cite{2020SciPy-NMeth}: \texttt{SpAdd} performs an element-wise addition ($X \oplus Y$) of the non-zero elements in $X$ and $Y$; the resulting values are converted to binary via the {\color{orange}\texttt{bool}} operator and saved in the \texttt{mask}, which corresponds to node indices of the union set $\mathcal{S}_Q$. 
    \item {\color{blue}\texttt{SpAdd}} are applied between $\texttt{mask}$ and each $\texttt{SpG}[u,:],\,\forall u\in Q$ following by the reduction {\color{red}`-1'}, which explicitly adds missing values (all zeros by default) to structural features $\mathcal{Z}_{u,x}$ for all $x$ if $x\not\in \mathcal{S}_u$ while $x\in \mathcal{S}_Q$.
    \item When the two-level indexing is enabled, the results of \texttt{SpJoin} are pointers saved in $\texttt{SFptr}$, which can be used to gather the values of structural features $\mathcal{Z}_Q$ from the array $D_{\text{SF}}$.
\end{enumerate}
Multithreading is employed to leverage the pattern of single program multiple data in arithmetic operations of \texttt{SpJoin}. Since the processing time of each query linearly depends on the size of $\mathcal{S}_Q$, we further divide queries of each training batch into groups with nearly balanced sums of $|\mathcal{S}_Q|$'s, and assign one thread per group to mitigate potential delays caused by uneven workloads.

\textbf{Comparison with SUREL~\cite{yin2022algorithm}} SUREL adopts a hash-based join operator to construct query-level structural features, but its overall computation and memory cost is much higher than \proj. This is due to the presence of numerous repeated nodes in walks, depicted in Fig. \ref{fig:subg_sampler} (b). The set-based input of \proj substantially reduces the workload of transferring data from CPU to GPU and also requires fewer per-query operations on GPU to process transmitted $\mathcal{Z}_Q$'s. As Table \ref{tab:store} shows, \proj reduces time complexity from $O(mM)$ to $O(K)$, where $K <mM$ is the average size of sampled node sets. These advantages ultimately enable \proj to achieve superior performance in terms of efficiency and scalability.

\subsection{Set Neural Encoders}
After joining node sets for each query $Q$, the resulting $(\mathcal{S}_Q, \mathcal{Z}_{Q})$ acts as the query-induced subgraph $\mathcal{G}_Q$ and then is fed into a neural encoder for prediction. The mini-batch training procedure of multiple queries is summarized in Algorithm \ref{alg:sgrl}. Next, we introduce neural encoders supported by \proj.

The adopted neural encoders are simple. For each $(\mathcal{S}_Q, \mathcal{Z}_{Q})$,
\begin{equation}
    h_Q = \texttt{AGGR}\left(\{enc(\mathcal{Z}_{Q,x})|x \in \mathcal{S}_Q\}\right) \in \RR^{d}.
    \label{eq:enc}
\end{equation}
Here, $enc(\cdot)$ encodes query-level structural features $\mathcal{Z}_{Q,x}$ using a multi-linear perception (MLP). If node attributes are present, they can be appended after structural features as $\mathcal{Z}_{Q,u} || X_u$. $\texttt{AGGR}$ is used to aggregate the encoded features, which can be any neural encoders applicable to sets such as mean/sum/max pooling or set transformers. Currently, \proj supports the implementations of $\texttt{AGGR}$ in mean pooling, LSTM~\cite{hamilton2017inductive}, and attention~\cite{velivckovic2017graph}. Note that, the LSTM applies random permutations to the elements in the set before encoding them as a sequence; while the attention first computes soft attention scores based on the output of $enc(\cdot)$ for each set element and then performs attention-score-weighted pooling. Sec. \ref{sec:param} empirically demonstrates that the choice of $\texttt{AGGR}$ has non-trivial effects on prediction performance. Lastly, a fully connected layer takes the readout $h_Q$ as input to make the final prediction $\hat{y}_Q$. In our experiments, all SGRL tasks are formulated as binary classification, and thus Binary Cross Entropy is used as the loss function $\mathcal{L}$.

\begin{algorithm}[tp]
\caption{\label{alg:sgrl} The mini-batch training pipeline of \proj} 
\KwIn{Given a graph $\mathcal{G}(\mathcal{V},\mathcal{E},X)$, a group of queries $\{(Q, y_Q)\}$ for training, batch size $B$, a set \texttt{SAMPLER}, a structure \texttt{ENCODER}, and a set \texttt{AGGR}}
\KwOut{A neural network for encoding subgraphs $enc(\cdot)$}
Preprocessing: \texttt{SAMPLER} and \texttt{ENCODER} $\to (\mathcal{S}_u, \mathcal{Z}_u)$ for all $u\in \mathcal{V}$; convert and save $(\mathcal{S}_u, \mathcal{Z}_u)$'s as \texttt{SpG} objects.\\
\For{each mini-batch $\mathcal{Q}_B=\{...,Q,...\}$}{
Generate negative training queries (if not given) $\{...,\bar{Q},...\}$ by random sampling and put them into $\mathcal{Q}_B$\;
Call \texttt{SpJoin} operator to perform joining on \texttt{SpG} objects $\{(\mathcal{S}_u, \mathcal{Z}_u)|u\in Q\}$ for all queries $Q\in\mathcal{Q}_B$ in parallel\;
Encode the joined results $(\mathcal{S}_Q, \mathcal{Z}_Q)$ as proxy of subgraphs via Eq. \eqref{eq:enc} with specified \texttt{AGGR} and get the prediction $\hat{y}_Q$ from readout $h_Q$ by multithreads\;
Backward propagation based on the loss $\mathcal{L}(\hat{y}_Q, y_Q)$.
}
\end{algorithm}
\begin{table}
\vspace{-9mm}
\end{table}

\section{Evaluation}
In this section, we aim to evaluate the following questions:
\begin{itemize}[leftmargin=*]
    \item Regarding space and time complexity, how much improvement can \proj achieve by adopting node sets instead of walks compared to the SOTA framework SUREL?
    \item Can \proj provide comparable prediction performance to all baselines using or not using subgraph-based methods?
    \item How sensitive is \proj to choices of different set samplers, structure encoders, and set neural encoders?
    \item How do sparse storage \texttt{SpG} and parallelism in \texttt{SpJoin} operator perform and benefit the overall performance of \proj?
\end{itemize}

\subsection{Experiment Setup}
Extensive experiments have been performed to evaluate \proj using nine homogeneous, heterogeneous, and higher-order homogeneous graphs on three types of tasks: link prediction, relation type prediction, and higher-order pattern prediction. A homogeneous graph is a graph that does not contain node/link types, while a heterogeneous graph includes various node/link types. In our setting, higher-order graphs are hypergraphs consisting of hyperedges connecting two or more nodes.

\textbf{Datasets} Table \ref{tab:data} summarizes the statistics of datasets used to benchmark SGRL methods. Five datasets are selected from the Open Graph Benchmark (OGB, \cite{hu2020open}) for link and relation type prediction, including social networks of citation - \texttt{citation2} and collaboration - \texttt{collab}; biological network of protein interaction - \texttt{ppa} and blood vessels - \texttt{vessel}; and one heterogeneous academic network \texttt{ogb-mag}, which contains node types of paper (P), author (A) and their extracted relations. The \texttt{vessel} dataset is a large ($>$3M nodes), sparse, biological graph recently constructed from mouse brains \cite{paetzold2021whole}, and has unique significance for examining GRL in scientific discovery. The structure of vessels illustrates the spatial organization of the brain’s microvasculature, which can be used for early detection of neurological disorders, e.g. Alzheimer’s and stroke.
Two hypergraph datasets collected by \cite{benson2018simplicial} are used for higher-order pattern prediction: \texttt{DBLP-coauthor} is a temporal hypergraph, where each hyperedge denotes a time-stamped paper connecting all its authors. \texttt{tags-math} contains groups of tags applied to questions on the website math.stackexchange.com as hyperedges. For higher-order pattern prediction tasks, the number of hyperedges is the main computation bottleneck, in which one may connect more than two nodes.
Two industry-level graphs, \texttt{criteo-click} with \emph{16.5M} records of online banner ads clicking~\cite{diemert2017attribution} and \texttt{twitter-2010} with \emph{1.5B} user following relations~\cite{kwak2010twitter} are used to examine the model scalability for real-world applications.

\textbf{Settings} For link prediction, OGB's standard data split is used to isolate validation and test links from the input graph. For prediction tasks of relation type and higher-order pattern, the same procedure to prepare graph data is adopted as in \cite{yin2022algorithm}: the relations of paper-author (P-A, "written by") and paper-paper (P-P, "cited by") are selected; higher-order queries in hypergraph datasets are node triplets, where the goal is to predict whether it will foster in a hyperedge given two of them have observed pairwise connections; to learn the representation on hypergraphs, we project hyperedges into cliques and treat the projection results as ordinary graphs. All experiments are run 10 times independently, and we report the mean performance and standard deviation.

\begin{table}[tp]
\centering
\caption{\small{Summary Statistics for Evaluation Datasets.}}
\label{tab:data}
\vspace{-4mm}
\begin{center}
  \resizebox{0.47\textwidth}{!}{
  \begin{tabular}{lcrrc}
    \toprule
    \textbf{Dataset}&\textbf{Type}&\textbf{\#Nodes}&\textbf{\#Edges}&\textbf{Split(\%)}\\
    \midrule
    \texttt{criteo-click} & Homo./Bipartite & \begin{tabular}[c]{@{}r@{}}Campaign: 675\\User: 6,142,256\end{tabular} & 16,468,027 & 97/1.5/1.5\\ \midrule
    \texttt{twitter-2010}  & Homo./Social. & 41,652,230 & 1,468,364,884 & 99.98/0.01/0.01\\
    \texttt{citation2} & Homo./Social. & 2,927,963 & 30,561,187 & 98/1/1\\
    \texttt{collab} & Homo./Social. & 235,868 & 1,285,465 &  92/4/4\\ \hline
    \texttt{ppa} & Homo./Bio. & 576,289 &  30,326,273 & 70/20/10\\
    \texttt{vessel} & Homo./Bio. & 3,538,495 & 5,345,897 & 80/10/10\\ \hline
    \texttt{ogb-mag} & Hetero. & \begin{tabular}[c]{@{}r@{}}(P): 736,389\\ (A): 1,134,649\end{tabular} & \begin{tabular}[c]{@{}r@{}}P-A: 7,145,660\\ P-P: 5,416,271\end{tabular} & 99/0.5/0.5\\ \hline
    \texttt{tags-math} & Higher. & 1,629 & \begin{tabular}[c]{@{}r@{}}projected: 91,685 \\ hyperedges: 822,059 \end{tabular} & 60/20/20\\
    \begin{tabular}[l]{@{}c@{}} \texttt{DBLP-} \\ \texttt{coauthor} \end{tabular} & Higher. & 1,924,991  & \begin{tabular}[c]{@{}r@{}}projected: 7,904,336 \\ hyperedges: 3,700,067  \end{tabular} &  60/20/20\\
  \bottomrule 
\end{tabular}}
\end{center}
\vspace{-6mm}
\end{table}

\begin{table*}[htp]
\caption{\small{Prediction Performance for Links, Relation Types and Higher-Order Patterns: the best (bold) and the second best (underlined).}}
\label{tab:ogb}
\centering
\resizebox{0.96\textwidth}{!}{
  \begin{tabular}{lccc|ccc|lcccc}
    \toprule
    \multirow{2}{*}{\textbf{Models}}  & \texttt{citation2} & \texttt{click} & \texttt{twitter} & \texttt{collab} & \texttt{ppa} & \texttt{vessel} & \multirow{2}{*}{\textbf{Models}} & \texttt{MAG(P-A)} & \texttt{MAG(P-P)} & \texttt{tags-math} & \texttt{DBLP-coauthor}\\
    & \multicolumn{3}{c|}{MRR (\%)} & Hits@50 (\%) & Hits@100 (\%) & ROC-AUC & & \multicolumn{4}{c}{MRR (\%)}  \\
    \midrule
    \textbf{GCN} & 84.74±0.21 & 5.31±0.17 & OOM & 44.75±1.07 & 18.67±1.32 & 43.53±9.61 & \textbf{H*GCN} & 39.43±0.29 & 57.43±0.30 & 51.64±0.27& 37.95±2.59 \\
    \textbf{GraphSAINT} & 79.85±0.40 & 2.86±0.63 & 4.12±0.73 & 53.12±0.52 & 3.83±1.33 & 47.14±6.83 & \textbf{H*SAGE} & 25.35±1.49 & 60.54±1.60 & 54.68±2.03 & 22.91±0.94\\
    \textbf{GDGNN} & 86.96±0.28 & 13.30±0.45 & \underline{49.86±0.39} &54.74±0.48 & 45.92±2.14 & 75.84±0.08 & \textbf{R-GCN} & 37.10±1.05 & 56.82±4.71  & - & - \\ 
    \textbf{SEAL} & 87.67±0.32 & OOM & OOM & \underline{63.64±0.71} & 48.80±3.16 & 80.50±0.21 & \textbf{SUREL} & \underline{45.33±2.94} & \textbf{82.47±0.26} & \underline{71.86±2.15} & \underline{97.66±2.89} \\
    \textbf{SUREL} & \textbf{89.74±0.18} & \underline{40.39±0.61} & OOM & 63.34±0.52 & \underline{53.23±1.03} & \textbf{86.16±0.39}  & \textbf{\proj} & \textbf{58.81±0.42} & \underline{80.45±0.13} & \textbf{77.73±0.16} & \textbf{99.83±0.02} \\
    \textbf{\proj} & \underline{88.90±0.06} & \textbf{60.87±0.15} & \textbf{55.67±0.67} & \textbf{64.10±1.06} & \textbf{54.32±0.44} & \underline{85.73±0.88} & / & / & / & / & / \\
  \bottomrule
\end{tabular}}
\end{table*}

\begin{table*}[tp]
\caption{\small{Breakdown of Runtime, Memory Consumption for Different Models on Prediction of Link, Relation Type, and Higher-order Pattern. The column Train records the runtime per 10K queries.}}
\label{tab:cost}
\centering
\resizebox{2.1\columnwidth}{!}{
\begin{tabular}{l|r|rr|rr|r|rr|rr|r|rr|rr|r|rr|rr}
\toprule
\multirow{2}{*}{\textbf{Models}} & \multicolumn{3}{c}{\textbf{Runtime (s)}}  & \multicolumn{2}{c|}{\textbf{Memory (GB)}} & \multicolumn{3}{c}{\textbf{Runtime (s)}}  & \multicolumn{2}{c|}{\textbf{Memory (GB)}} & \multicolumn{3}{c}{\textbf{Runtime (s)}}  & \multicolumn{2}{c|}{\textbf{Memory (GB)}} & \multicolumn{3}{c}{\textbf{Runtime (s)}}  & \multicolumn{2}{c}{\textbf{Memory (GB)}}\\ \cmidrule(r{0.5em}){2-21}
& Prep. & Train & Inf. & RAM & SDRAM & Prep. & Train & Inf. & RAM & SDRAM & Prep. & Train & Inf. & RAM & SDRAM & Prep. & Train & Inf. & RAM & SDRAM\\ \midrule
\texttt{Dataset} & \multicolumn{5}{c|}{\texttt{criteo-click}} & \multicolumn{5}{c|}{\texttt{twitter-2010}} & \multicolumn{5}{c|}{\texttt{citation2}} & \multicolumn{5}{c}{\texttt{ppa}}\\ \midrule
    \textbf{GCN}  & 3 & 0.085 & 8 & 3.1 & 62.74 & - & - & - & - & OOM  & 17 & 21.74 & 105 & 9.3 & 36.84 & 2 & 0.026 & 1.2 & 4.6 & 11.35 \\
    \textbf{GraphSAINT} & 1 & 0.012 & 20 & 13.1 & 8.79 & 111 & 0.009 & 920 & 253 & 76.60 & 151 & 1.79 & 107 & 9.6 & 9.78 & 10 & 0.003 & 1.5 & 4.9 & 23.06\\
    \midrule
    \textbf{GDGNN}  & 215 & 1.43 & 2,928 & 16.2 & 23.77 & 1204 & 1.84 & 9,744 & 188 & 79.34 & 338 & 2.26 & 5,460 & 40.6 & 16.96 & 127 & 1.77 & 902 & 21.1 & 10.27\\
    \textbf{SEAL} & - & - & - & OOM & - & - & - & - & OOM & - & 46 & 3.52 & 24,626 & 35.4 & 5.71 & 46 & 10.57 & 3,988 & 9.5 & 12.13\\
    \textbf{SUREL} & 2 & 1.59 & 2,307 & 11.7 & 16.25 & - & - & - & OOM & - & 151 & 4.14 & 6,081 & 25.1 & 9.68 & 31 & 2.68 & 1,429 & 13.6 & 31.01\\
    \textbf{\proj} & 22 & 0.23 & 502 & 10.4 & 11.93 & 327 & 0.26 & 3,779 & 210 & 49.44 & 130 & 0.35 & 1,389 & 16.7 & 4.75 & 69 & 0.72 & 201 & 9.8 & 19.02 \\
\bottomrule
\end{tabular}}
\end{table*}

\textbf{Baselines} We consider two classes of baselines. \textit{Canonical GNNs}: GCN \cite{kipf2016semi}, GraphSAGE \cite{hamilton2017inductive}, GraphSAINT \cite{zeng2019graphsaint} and their variants with the prefix `H*' that are directly applied for heterogeneous graphs with node types and for hypergraphs through clique expansion. R-GCN \cite{schlichtkrull2018modeling} performs relational message passing on heterogeneous graphs. \textit{SGRL Models}: SEAL \cite{zhang2018link,zhang2021labeling}, GDGNN \cite{kong2022geodesic}, and SUREL \cite{yin2022algorithm}. SEAL adopts online subgraph sampling due to its intractable space needs for offline extraction. Fig. \ref{fig:subg_sampler} (a) compares the time cost for subgraph sampling across different SGRL methods. We use all baselines' official implementations with tuned hyperparameters to match their reported results.

\label{sec:exp}
\textbf{Hyperparameters} By default, \proj uses the walk-based sampler, the structural encoder LP, and the better set neural encoder tuned between mean pooling and attention. \proj adopts a 2-layer MLP as $enc(\cdot)$ in Eq.~\eqref{eq:enc} followed by a 2-layer MLP classifier to map set-aggregated readouts for final predictions. Default training hyperparameters: learning rate \texttt{lr=1e-3} with early stopping of 5 epochs, dropout \texttt{p=0.1}, Adam \cite{kingma2014adam} as the optimizer. Analysis of parameters $M$ and $m$ to control the walk-based sampler and $K$ to control the metric-based sampler and selection of structure encoders and set neural encoders are studied in Sec. \ref{sec:param}.

\textbf{Evaluation Metrics} The evaluation metrics include Hits@P, Mean Reciprocal Rank (MRR), and Area Under Curve (ROC-AUC). Hit@P counts the ratio of positive samples ranked at the top-P place against negative ones. MRR first computes the inverse of the rank of the first correct prediction and then takes the average of obtained reciprocal ranks for a sample of queries. For all datasets adopting MRR, each positive query is paired with 1000 randomly sampled negative test queries, except \texttt{tags-math} using 100 and \texttt{crieo-click} using 650. ROC-AUC follows the standard definition to measure the model's performance in binary classification.

\textbf{Environment} We use a server with two Intel Xeon Gold 6248R CPUs, 512GB DRAM, and NVIDIA A100 (80GB) GPU. \proj is built on PyTorch 1.12 and PyG 2.2. Set samplers are implemented in C, OpenMP, NumPy, Numba, and \texttt{uhash}, integrated into Python scripts; \texttt{SpG} is customized based on the CSR format of Scipy \cite{2020SciPy-NMeth}.

\subsection{Prediction Accuracy Comparison}
Table \ref{tab:ogb} shows the prediction performance of different methods. SGRL models significantly outperform canonical GNNs on these six link prediction benchmarks, especially on two challenging biological datasets \texttt{ppa} and \texttt{vessel}. Predicting links in biological datasets requires richer structural information that canonical GNNs have limited expressive power to capture. Within SGRL models, \proj achieves comparable performance to SUREL and outperforms SEAL, which validates the effectiveness of the proposed set-based representation for subgraphs. For predictions of relation type and higher-order pattern, we observe additional performance gains (+2$\sim$13\%) from \proj compared to SUREL on three of the four datasets. A large performance gap exists between canonical GNNs and SGRL models, particularly in the higher-order case. This demonstrates the inherent limitations of canonical GNNs to make predictions of complex relations involving multiple nodes.

\subsection{Efficiency and Scalability Analysis}
\paragraph{Improved Efficiency in Training and Inference.}
Table \ref{tab:cost} compares model runtime and memory usage on the four largest benchmarks. \proj offers a reasonable training time compared with canonical GNNs. It shows clear improvement in inference compared to the current SOTA framework SUREL (3-11$\times$ speedups across all datasets) and its predecessor SEAL ($\sim$20$\times$ speedups). \proj achieves comparable and even lower RAM usage than canonical GNNs. Compared to other SGRL models, it can save up to half of RAM with lower usage of GPU SDRAM. This is attributed to set-based subgraphs eliminating node duplicates with structural features, which is further echoed by the analysis in Table \ref{tab:store} and the empirical results in Table \ref{tab:cost}. The key factor scales \proj to billion-size graphs is its set-based subgraph with the sparse design, while GCN (full adjacency matrix), SEAL (complex subgraph extraction), and SUREL (dense walks with duplicate nodes) are all out of memory (OOM) on \texttt{twitter-2010}.

\begin{figure}[tp]
\centering
\begin{subfigure}[b]{0.23\textwidth}
    \centering
    \includegraphics[height=0.65\textwidth]{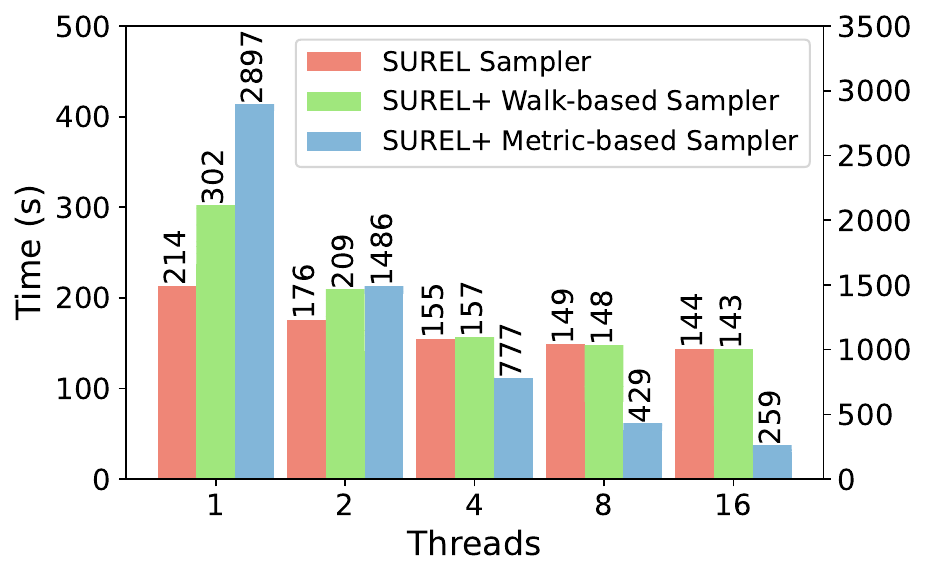}
    \vspace{-5mm}
    \caption{Runtime \label{fig:runtime}}
\end{subfigure}
\hfill
\begin{subfigure}[b]{0.23\textwidth}
    \centering
    \includegraphics[height=0.65\textwidth]{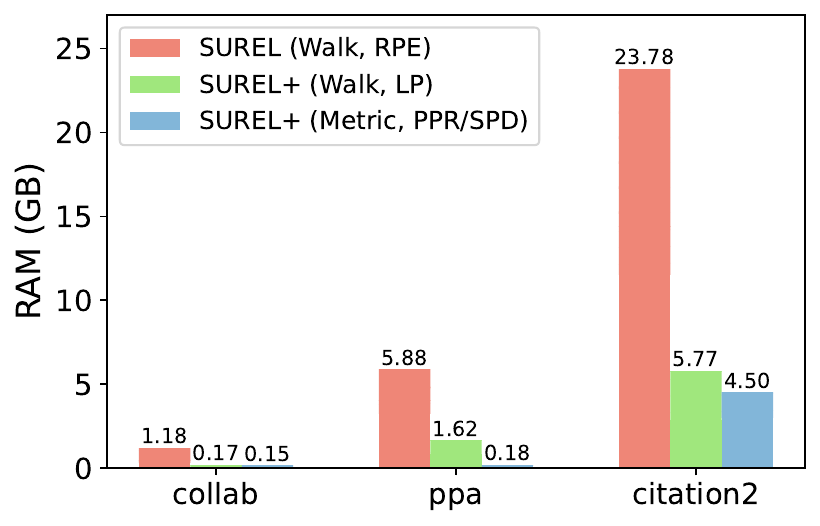}
    \vspace{-5mm}
    \caption{Memory\label{fig:storage}}
\end{subfigure}
\vspace{-4mm}
\caption{\small{Comparison of Runtime, Memory Consumption across Different Offline Processing Strategies (the walk-based sampler: $m=4,M=200$, the metric-based sampler: $K=150$). The highlighted areas break down the total consumption w.r.t. (a) sampling, structure encoding, sparse object construction; (b) structural features, node indices/pointers, and sampled walks (SUREL sampler only). \label{fig:compare}}}
\vspace{-2mm}
\end{figure}

\paragraph{Profiling Different Strategies for Offline Processing} Fig. \ref{fig:runtime} reports the time cost of different samplers with multithreading on \texttt{citation2}. Fig. \ref{fig:storage} shows memory consumption to store different types of sampled data (walks in SUREL \cite{yin2022algorithm} or sets in \proj) and associated structural features (LPs, SPDs, PPR scores). Compared to the SUREL sampler, the walk-based sampler in \proj is more efficient and only adds one extra minute for encoding and converting data to \texttt{SpG} format (slash/dash marked in Fig. \ref{fig:runtime}), while achieving $6.94\times$, $3.63\times$ and $4.12\times$ memory savings on three OGB datasets, respectively. Those savings are crucial for model scalability as they reduce data transfer from CPU to GPU and reduce GPU operations on duplicate nodes. These two factors dominate the online stage and thus lead to improved memory usage and runtime of \proj in Table \ref{tab:cost}. In addition, the PPR-based sampler has better scaling performance with more threads. When PPR scores or SPDs are used as structural features, \proj further reduces the memory footprint, though they often slightly harm prediction performance.

Note that, in the above comparison of memory cost, techniques of compressing structural features are adopted both in SUREL (locally) and \proj (globally). When LPs are used as structural features, the two-level indexing in \texttt{SpG} achieves compression of $493\times$, $11318\times$, $19527\times$ on three datasets listed in Fig. \ref{fig:storage}.

\begin{figure}
\begin{subfigure}[b]{0.23\textwidth}
    \centering
    \includegraphics[height=0.67\textwidth]{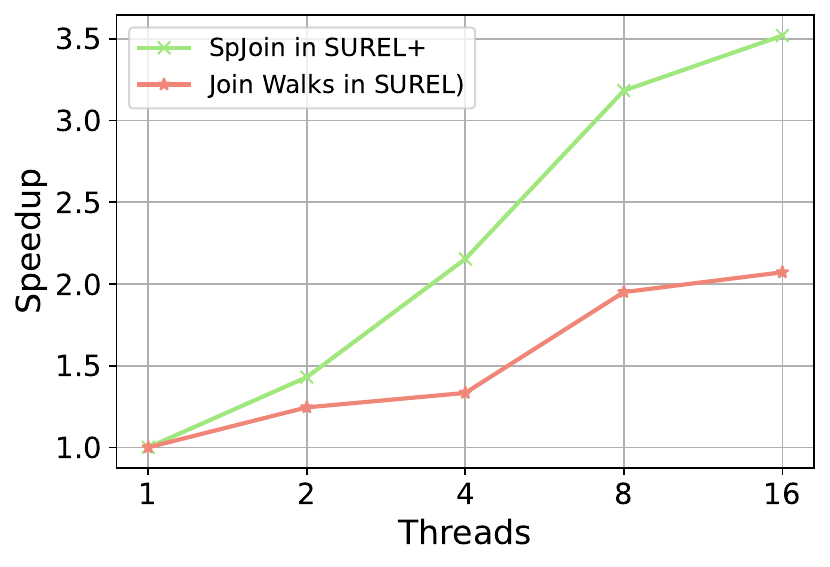}
    \vspace{-2mm}
    \caption{Speedup\label{fig:scaling}}
\end{subfigure}
\hfill
\begin{subfigure}[b]{0.23\textwidth}
    \centering
    \includegraphics[height=0.67\textwidth]{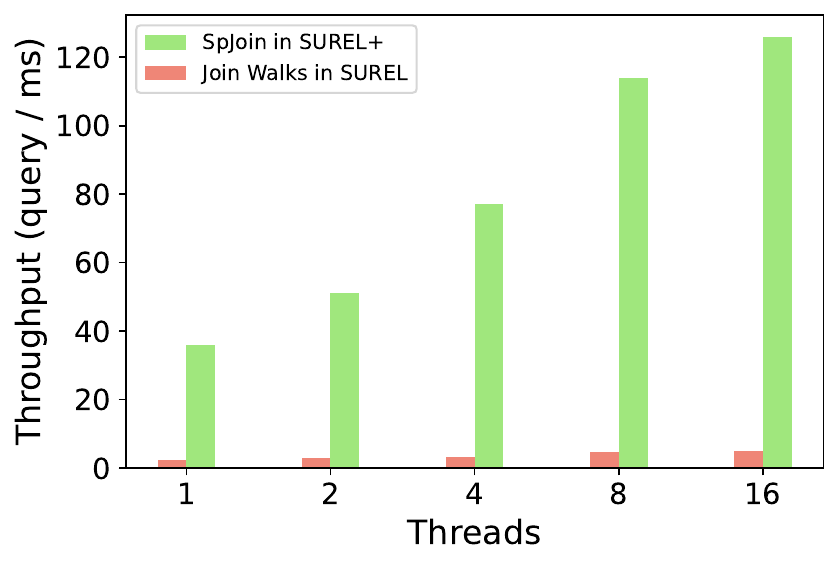}
    \vspace{-6mm}
    \caption{Throughput}
\end{subfigure}
\vspace{-4mm}
\caption{\small{Scaling Performance Comparison of \texttt{SpJoin} in \proj (with average set size $\bar{|S_u|}=351$) and Join Walks in SUREL (with walk size $m=4,M=200$) against Different Numbers of Threads.}\label{fig:join}}
\vspace{-4mm}
\end{figure}

\paragraph{Scaling Analysis for \texttt{SpJoin}} Fig. \ref{fig:join} shows the speedups and throughput of the \texttt{SpJoin} operator for constructing query-level structural features via multithreading, where the walk join operation of SUREL is used for comparison. SUREL employs a hash-based search for joining walks, which has unfavorable memory access patterns and suffers from imbalanced workloads due to inconsistent searching times across different threads. \proj gains more benefits from multithreading, thanks to sparse arithmetic operations and batch-wise load balancing used in \texttt{SpJoin}. 

\subsection{Comparison between Different Set Samplers, Structural Features and Set Neural Encoders \label{sec:param}}
\proj is a modularized framework that supports different set samplers (walk- and metric-based), structural features (LP, SPD, PPR), and set neural encoders \texttt{AGGR} (mean pooling, LSTM, attention). 

Table \ref{tab:param} shows the prediction performance and inference runtime by adopting different combinations of structure encoders and set neural encoders. Landing probabilities (LPs) as structural features perform the best on all three OGB datasets while being the slowest for inference. By recording the landing probabilities over different steps of walks, LPs provide structural information in finer granularity than scalar values of SPDs and PPR scores. Furthermore, the adopted link prediction task might favor more local information held by LPs and SPDs than global information carried by PPR scores. The authors conjecture that other tasks that rely on more global information may favor PPR scores. In comparison, no set neural encoder is always a winner. Attention seems to perform the best on average while slower than mean pooling. LSTM is the slowest. On the two social networks (\texttt{citation2} and \texttt{collab}), mean pooling can provide comparable prediction results with much fewer parameters. However, prediction on the biological network (\texttt{ppa}) requires more expressive and complicated encoders, where LSTM and attention are favored as they can model more complex interactions between sampled nodes in the union set $\mathcal{S}_Q$.

\begin{figure}[t]
\centering
\begin{subfigure}[b]{0.15\textwidth}
    \centering
    \includegraphics[height=0.75\textwidth]{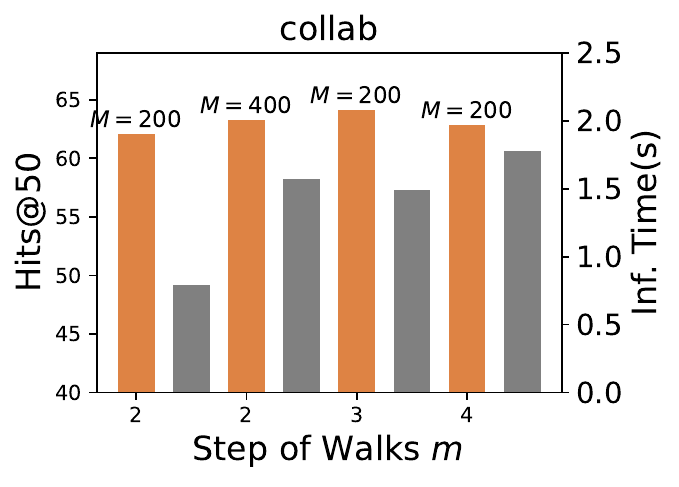}
    \vspace{-1mm}
\end{subfigure}
\hfill
\begin{subfigure}[b]{0.15\textwidth}
    \centering
    \includegraphics[height=0.75\textwidth]{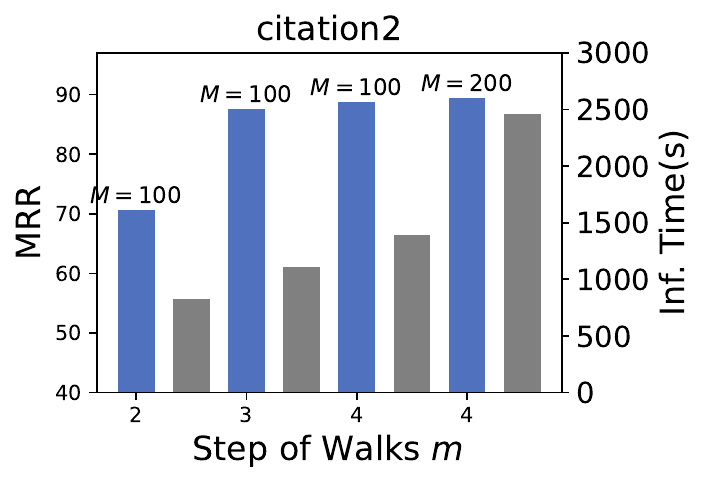}
    \vspace{-1mm}
\end{subfigure}
\hfill
\begin{subfigure}[b]{0.15\textwidth}
    \centering
    \includegraphics[height=0.72\textwidth]{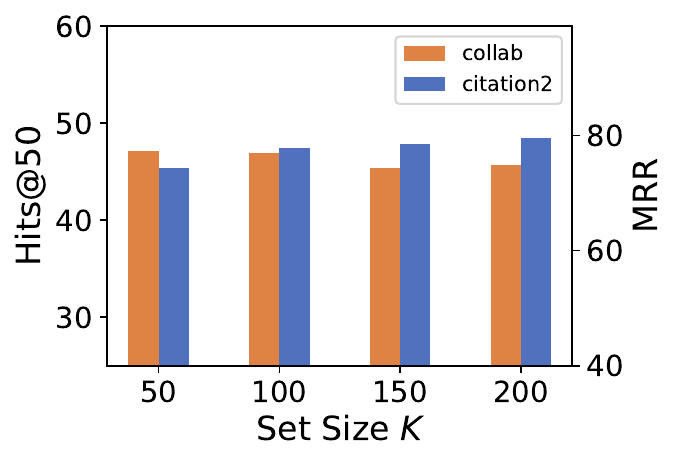}
    \vspace{-1mm}
\end{subfigure}
\vspace{-6mm}
\caption{\small{Hyperparameter Analysis of Set Samplers (Prediction Performance v.s. Time Cost). Walk-based: the number $M$ and the step $m$ of walks, LPs as structural features; Metric-based: the set size $K$, PPR scores as structural features. \label{fig:param}}}
\vspace{-4mm}
\end{figure}

Fig. \ref{fig:param} compares prediction results and inference time by using different hyperparameters $m, M$, and $K$ of set samplers, which heavily affects the coverage of sampled neighborhoods and computation overhead. The performance consistently increases if the walk-based sampler uses a larger $M$, but is not guaranteed for a larger $m$ (broader exploration). Better coverage with a larger $K$ is usually beneficial for the metric-based sampler over \texttt{citation2} but not for \texttt{collab}, which is due to different characteristics of these two datasets and is also observed by \cite{yin2022algorithm}. In general, small sampling parameters $m~(2\sim4), M~(100\sim400)$ and $K~(50\sim200)$ can yield satisfactory performance with fast inference speed that achieves the trade-off between accuracy and efficiency.

\begin{table}[tp]
\caption{\small{Prediction Performance and Inference Time of \proj with Different Combinations of Structure Features (LP, SPD, PPR) and Set Neural Encoders (Mean, LSTM, Attn.). The best and the second best are highlighted in bold and underlined accordingly.}}
\vspace{-2mm}
\label{tab:param}
\begin{center}
\resizebox{\columnwidth}{!}{
  \begin{tabular}{l|ccccc}
    \toprule
    \textbf{Dataset} & PPR+Mean & SPD+Mean & LP+Mean & LP+LSTM & LP+Attention\\
    \midrule
    \multirow{2}{*}{\texttt{citation2}} & 78.59±0.38 & 87.99±1.07 & \underline{88.55±0.15} & 88.46±0.34 & \textbf{88.90±0.06}\\
    & \textbf{834} & \underline{1057s} & 1389s & 3678s & 2171s\\
    \multirow{2}{*}{\texttt{collab}} & 47.15±0.21 & 62.11±0.13 & \textbf{64.10±1.06} & 61.31±1.37 & \underline{62.85±1.19}\\
    & \textbf{1.4s} & \underline{1.7s} & 2.0s & 3.5s & 2.3s\\
    \multirow{2}{*}{\texttt{ppa}} & 13.28±1.20 & 41.06±1.70 & 46.41±1.65 & \textbf{54.45±1.35} & \underline{54.32±0.44}\\
    & \textbf{63s} & \underline{126s} & 165s & 322s & 201s\\
  \bottomrule
\end{tabular}}
\end{center}
\vspace{-2mm}
\end{table}

\section{Conclusion}
This work proposes a novel framework \proj for scalable subgraph-based graph representation learning. \proj avoids costly subgraph extraction by decoupling it into sampled node sets with structural features, whose join can function as query-induced subgraphs for prediction. 
\proj benefits from the reusability and compactness of pre-sampled node sets across different queries. Compared to the SOTA framework SUREL, the set-based subgraph of \proj substantially reduces space and time complexity by avoiding heavy node duplication in sampled walks. To handle irregularly sized node sets, \proj designs a customized sparse storage \texttt{SpG} and a sparse join operator \texttt{SpJoin}, providing memory-efficient storage with fast access. In addition, \proj adopts a modular design, enabling users to choose different set samplers, structure encoders, and set neural encoders flexibly based on the nature of their SGRL tasks. Extensive experiments on three types of prediction tasks over nine real-world graph benchmarks show that \proj significantly improves scalability, memory efficiency, and prediction accuracy compared to current SGRL methods and canonical GNNs.

\begin{acks}
The authors would like to thank Rongzhe Wei and Yanbang Wang for their helpful discussions and valuable feedback. Haoteng Yin and Pan Li are supported by the 2021 JPMorgan Faculty Award, NSF awards OAC-2117997, IIS-2239565.
\end{acks}

\bibliographystyle{ACM-Reference-Format}
\bibliography{ref}

\appendix
\section{Notations}\label{apd:notation}
Frequently used symbols are summarized in Table \ref{tab:notation}.

\section{More Details}
\begin{table}
\centering
\caption{Summary of Frequently Used Notations.}
\vspace{-2mm}
\label{tab:notation}
\begin{center}
  \begin{tabular}{lp{6.5cm}}
    \toprule
    \textbf{Symbol}&\textbf{Meaning}\\
    \midrule
    Q & a query (set of nodes), i.e. $Q=\{u,v,w\}$\\
    $\mathcal{Q}$ & a set of queries, i.e. $Q\in \mathcal{Q}$\\
    $\mathcal{G}_u$ & a subgraph induced by node $u$\\
    $\mathcal{G}_Q$ & a subgraph induced by query $Q$\\
    $\mathcal{S}_u$ & a set of unique nodes sampled from the neighborhood of the seed node $u$ \\
    $\mathcal{Z}_{u,x}$ & structural features of node $x$ regarding the seed node $u$ (all zeros if $x \notin \mathcal{S}_u$)\\
    $\mathcal{Z}_u$ & collection of structural features for all nodes in $\mathcal{S}_u$ as $\mathcal{Z}_u = \{\mathcal{Z}_{u,x}|x\in \mathcal{S}_u\}$\\
    $||$ & the concatenation that joins node-level structural features, i.e. join $\mathcal{Z}_{\cdot,x}$ for a query $Q=\{u,v,w\}$ as $[\mathcal{Z}_{u,x},\mathcal{Z}_{v,x},\mathcal{Z}_{w,x}]$.\\
    $\mathcal{Z}_{Q,x}$ & query-level structural features for node $x$ regarding the query $Q$, $\mathcal{Z}_{Q,x}=||_{u \in Q} \mathcal{Z}_{u,x}$\\
  \bottomrule
\end{tabular}
\end{center}
\vspace{-4mm}
\end{table}

\subsection{Other Related Works}
\textbf{Scalable GNN Design.} GNNs are the most widely used toolbox for graph representation learning nowadays, although they face certain challenges when directly applied to subgraph-based methods. To address the scalability of GNNs, current studies focus on improving graph subsampling and mini-batch training techniques~\cite{chiang2019cluster,zeng2019graphsaint}. However, graph subsampling used in GNNs fundamentally differs from subgraph extractions in SGRL. The goal of subsampling is to handle GPU memory overflow during full-batch training of GNN models. For SGRL, subgraphs sampled around a query serve as features for making predictions. Consequently, the scaling techniques developed for GNNs cannot be directly applied to SGRL. Another direction is to deploy distributed GNN systems for industry-level graphs. Unfortunately, these specialized techniques, including pipelining~\cite{wan2022pipegcn}, partitioned parallelism~\cite{wan2022bns}, and update with staleness~\cite{peng2022sancus} do not address the main bottleneck of subgraph extraction for SGRL methods.

\subsection{Model Design}
\textbf{Benefits of Subgraph-based Graph Representation Learning} First, subgraph-based representation is versatile for different types of tasks, especially when queries of certain tasks involving multiple nodes and relations, e.g. existence of a link, property of a motif, development of higher-order patterns; while canonical GNNs are limited to handle such polyadic dynamics via node-wise representations \cite{srinivasan2019equivalence,wang2021glass}. Second, subgraph-based models are more expressive by pairing with structural features to obtain most expressive structural representations \cite{srinivasan2019equivalence,li2020distance,wang2021glass,bouritsas2022improving}. However, canonical GNNs cannot capture intra-distance information and joint relations over multiple nodes, which are critical to distinguishing nodes in structural symmetry and making predictions over them (also refers to the example in Fig. \ref{fig:sgrl}). Lastly, subgraph-based methods decouple the model depth from the receptive field since extracted subgraphs are localized to certain hops: when adding more layers for non-linearity, it does not contaminate embedding with irrelevant nodes or get over-smooth as canonical GNNs do. This results in a more robust representation and is particularly beneficial for modeling relations beyond singleton.

\begin{table*}[htp]
\centering
\caption{Summary Statistics and Experimental Setup for Evaluation Datasets.}
\vspace{-3mm}
\label{tab:data-full}
  \resizebox{0.98\textwidth}{!}{
  \begin{tabular}{lcccccccc}
  \toprule\textbf{Dataset}&\textbf{Type}&\textbf{\#Nodes}&\textbf{\#Edges}&\textbf{Avg. Node Deg.}&\textbf{Density}&\textbf{Split Ratio}&\textbf{Split Type}&\textbf{Metric}  \\
    \midrule
    \texttt{criteo-click} & Homo./Bipartite & \begin{tabular}[c]{@{}r@{}}Campaign(C): 675\\User(U): 6,142,256\end{tabular} & 16,468,027 & 2.68 & N/A & 97/1.5/1.5 & Time & MRR\\\midrule
    \texttt{twitter-2010}  & Homo./Social. & 41,652,230 & 1,468,364,884 & 35.25 & 0.00017\% & 99.98/0.01/0.01 & Random & MRR\\
    \texttt{citation2} & Homo./Social. & 2,927,963 & 30,561,187 & 20.7 & 0.00036\% & 98/1/1 & Time & MRR\\
    \texttt{collab} & Homo./Social. & 235,868 & 1,285,465 & 8.2 & 0.0046\% & 92/4/4 & Time & Hits@50\\\midrule
    \texttt{ppa} & Homo./Bio. & 576,289 &  30,326,273 & 73.7 & 0.018\% & 70/20/10 & Throughput & Hits@100\\
    \texttt{vessel} & Homo./Bio. & 3,538,495 & 5,345,897  & 3.02 & 0.000085\% & 80/10/10 & Random & AUC-ROC\\\midrule
    \texttt{ogb-mag} & Hetero. & \begin{tabular}[c]{@{}l@{}}Paper(P): 736,389\\ Author(A): 1,134,649\end{tabular} & \begin{tabular}[c]{@{}l@{}}P-A: 7,145,660\\ P-P: 5,416,271\end{tabular} & 21.7 & N/A & 99/0.5/0.5 & Time & MRR\\\midrule
    \texttt{tags-math} & Higher. & 1,629 & \begin{tabular}[c]{@{}l@{}}91,685 (projected) \\ 822,059 (hyperedges) \end{tabular} & N/A & N/A & 60/20/20 & Time & MRR\\
    \texttt{DBLP-coauthor} & Higher. & 1,924,991  & \begin{tabular}[c]{@{}l@{}}7,904,336 (projected) \\ 3,700,067 (hyperedges) \end{tabular} & N/A & N/A & 60/20/20 & Time & MRR\\
  \bottomrule
\end{tabular}}
\vspace{-2mm}
\end{table*}

\begin{table*}[tp]
\caption{[Extended] Breakdown of Runtime, Memory Consumption for Different Models on Prediction of Link, Relation Type, and Higher-order Pattern. The column Train records the runtime per 10K queries.}
\vspace{-3mm}
\label{tab:cost_full}
\begin{center}
\resizebox{2.1\columnwidth}{!}{
\begin{tabular}{l|r|rr|rr|r|rr|rr|r|rr|rr|r|rr|rr}
\toprule
\multirow{2}{*}{\textbf{Models}} & \multicolumn{3}{c}{\textbf{Runtime (s)}}  & \multicolumn{2}{c|}{\textbf{Memory (GB)}} & \multicolumn{3}{c}{\textbf{Runtime (s)}}  & \multicolumn{2}{c|}{\textbf{Memory (GB)}} & \multicolumn{3}{c}{\textbf{Runtime (s)}}  & \multicolumn{2}{c|}{\textbf{Memory (GB)}} & \multicolumn{3}{c}{\textbf{Runtime (s)}}  & \multicolumn{2}{c}{\textbf{Memory (GB)}}\\ \cmidrule(r{0.5em}){2-21}
& Prep. & Train & Inf. & RAM & SDRAM & Prep. & Train & Inf. & RAM & SDRAM & Prep. & Train & Inf. & RAM & SDRAM & Prep. & Train & Inf. & RAM & SDRAM\\ \midrule
\texttt{Dataset} & \multicolumn{5}{c|}{\texttt{citation2}} & \multicolumn{5}{c|}{\texttt{ppa}} & \multicolumn{5}{c|}{\texttt{collab}} & \multicolumn{5}{c}{\texttt{vessel}}\\
\midrule
\textbf{GCN} & 17 & 21.74 & 105 & 9.3 & 36.84 & 2 & 0.026 & 1.2 & 4.6 & 11.35 & 2 & 0.005 & 0.05 & 2.5 & 5.50 & 5 & 0.076 & 0.3 & 2.8 & 36.98\\
\textbf{GraphSAINT} & 151 & 1.79 & 107 & 9.6 & 9.78 & 10 & 0.003 & 1.5 & 4.9 & 23.06 & 1 & 0.004 & 0.08 & 2.5 & 8.11 & 5 & 0.008 & 15 & 6.9 & 10.21\\ \midrule
\textbf{GDGNN} & 338 & 2.26 & 5,460 & 40.6 & 16.96 & 127 & 1.77 & 902 & 21.1 & 10.27 & 14 & 0.74 & 15 & 4.3 & 1.08 & 25 & 0.85 & 84 & 7.2 & 8.03\\
\textbf{SEAL} & 46 & 3.52 & 24,626 & 35.4 & 5.71 & 46 & 10.57 & 3,988 & 9.5 & 12.13 & 5 & 4.05 & 37 & 4.0 & 6.20 & 6 & 10.69 & 998 & 6.2 & 2.46\\
\textbf{SUREL} & 151 & 4.14 & 6,081 & 25.1 & 9.68 & 31 & 2.68 & 1,429 & 13.6 & 31.01 & 1 & 2.13 & 17 & 3.4 & 9.86 & 5 & 1.57 & 32 & 5.8 & 5.18\\
\textbf{\proj} & 130 & 0.35 & 1,389 & 16.7 & 4.75 & 69 & 0.72 & 201 & 9.8 & 19.02 & 7 & 0.27 & 2 & 2.8 & 3.37 & 3 & 0.31 & 3 & 3.3 & 1.25\\
\midrule
\texttt{Dataset} & \multicolumn{5}{c|}{\texttt{MAG(P-A)}} & \multicolumn{5}{c|}{\texttt{MAG(P-P)}} & \multicolumn{5}{c|}{\texttt{tags-math}} & \multicolumn{5}{c}{\texttt{DBLP-coauthor}}\\
\midrule
\textbf{H*GCN} & 3 & 0.03 & 9 & 5.0 & 21.56 & 4 & 0.03 & 13 & 5.5 & 21.66 & 2 & 0.004 & 1.3 & 2.4 & 3.10 & - & 0.58 & 95 & 8.0 & 25.80\\
\textbf{H*SAGE} & 3 & 0.03 & 10 & 5.0 & 20.29 & 4 & 0.03 & 13 & 5.5 & 20.28 & 1 & 0.003 & 1.3 & 2.4 & 3.10 & - & 0.32 & 77 & 7.5 & 24.70 \\
\textbf{R-GCN} & 1 & 0.52 & 5 & 5.3 & 26.34 & 1 & 0.52 & 4 & 5.1 & 31.41 &- & - & - & - & - & - & - & - & - & - \\ \midrule
\textbf{SUREL} & 10 & 3.20 & 1,998 & 7.3 & 7.18 & 15 & 0.99 & 1924 & 8.1 & 16.66 & - & 2.13 & 341 & 3.0 & 5.95 & 11 & 1.29  & 949 & 9.7 & 7.79 \\
\textbf{\proj} & 58 & 0.33 & 101 & 7.2 & 2.95 & 77 & 0.13 & 168 & 8.1 & 13.49 & 1 & 0.67 & 116 & 2.4 & 5.70 & 8 & 0.24 & 315 & 3.8 & 3.16\\
\bottomrule
\end{tabular}}
\end{center}
\vspace{-4mm}
\end{table*}

\subsection{Datasets}
The full statistics of benchmark datasets are summarized in Table \ref{tab:data-full}. OGB datasets\footnote{\url{https://ogb.stanford.edu/docs/dataset_overview/}} are selected to benchmark our proposed framework and other baselines. The benchmark contains large-scale graphs (millions of nodes/edges) for real-world applications (e.g., academic and biological networks) and provides standard, open-sourced evaluation metrics and toolkits. Note that, \texttt{vessel} is a newly added benchmark of a biological graph, with $>3M$ nodes and sparse vessel structures extracted from the whole mouse brain \cite{paetzold2021whole}, where nodes represent bifurcation points, and edges represent the blood vessels. Each node is associated with features of its physical location in the coordinate space $(x, y, z)$. The introduction of \texttt{vessel} provides a unique opportunity to examine graph representation learning approaches in neuroscience, especially in scaling subgraph-based methods to handle sparse and spatial graphs with millions of nodes and edges for scientific discovery. 

\texttt{criteo-click} contains a sample of 30 days of Criteo live traffic data, each corresponding to one impression (a banner) displayed to a user and whether it is clicked~\cite{diemert2017attribution}. Each record has 9 contextual features that are aggregated into a 270-dimensional edge feature. There are 675 unique campaign banners and 6.1M users, consisting of a bipartite graph of 16.5M edges: 97\% is used for training, and the rest is evenly split for validation and testing based on temporal orders. The task is to predict which campaign the user is most likely to click among 651 candidates. \texttt{twitter-2010} is an industry-level social network with \emph{1.5B} user following relations~\cite{kwak2010twitter}. An edge $(i,j)$ of this network indicates that user $i$ is followed by user $j$. 1\% of Twitter users who follow 10 to 1000 accounts are randomly sampled for evaluation. The task is to recommend which account they will most likely follow among 1001 candidates. The OGB formatted files of these two datasets are accessible via Box at \url{https://purdue.box.com/v/SGRL-LSC-dataset}.

\subsection{Baselines}
For link prediction and relation type prediction, baseline models are selected based on their scalability and prediction performance from the current OGB leaderboard \footnote{\url{https://ogb.stanford.edu/docs/leader_linkprop/}}. All models listed on the leaderboard are publicly accessible. We adopt their reported numbers on the leaderboard with verification. For the rest of the baselines, we benchmark these models using their official implementations with tuned hyperparameters as listed below.
\begin{itemize}[leftmargin=*]
    \item \textbf{Canonical GNNs}: a graph auto-encoder model that uses graph convolution layers to learn node-wise representations, including GCN~\citep{kipf2016semi}, GraphSAGE~\citep{hamilton2017inductive}, and their more scalable variants by employing graph subsampling, such as GraphSAINT~\citep{zeng2019graphsaint}.
    \item \textbf{R-GCN}\footnote{\url{https://github.com/pyg-team/pytorch\_geometric/blob/master/examples}}~\citep{schlichtkrull2018modeling}: a relational GCN that models heterogeneous graphs with different types of node/link.
    \item \textbf{SEAL}\footnote{\url{https://github.com/facebookresearch/SEAL_OGB}}~\citep{zhang2018link}: apply GCN on query-induced subgraphs attached with double radius node labeling to obtain subgraph-level readout for link prediction. SEAL shows great empirical performance on multiple graph machine learning benchmarks and promotes the deployment of subgraph-based models for scientific discovery. The implementation we tested is specialized for OGB datasets provided in \cite{zhang2021labeling}.
    \item \textbf{GDGNN}\footnote{\url{https://github.com/woodcutter1998/gdgnn}}~\citep{kong2022geodesic}: a subgraph-based model aggregates node representations generated by GNNs along geodesic paths between queried nodes for fast inference.
    \item \textbf{SUREL}\footnote{\url{https://github.com/Graph-COM/SUREL}}\cite{yin2022algorithm}: a walk-based computation framework to accelerate subgraph-based methods, where subgraphs are decomposed to pre-sampled walks and then are joined online to substitute the query-induced subgraph for prediction. By adopting the walk-based representation, SUREL achieves state-of-the-art scalability and prediction accuracy on SGRL tasks.
\end{itemize}

All canonical GNN baselines\footnote{\url{https://github.com/snap-stanford/ogb/tree/master/examples/linkproppred}} come with three \texttt{GCNConv}/\texttt{SAGEConv} layers of 256 hidden dimensions, and a tuned dropout ratio in $\{0,0.5\}$ for full-batch training. Canonical GNNs aggregate all node embeddings involved in a query as the representation of link/hyperedge, which is later fed into an MLP classifier for final prediction. In addition, all GNN models need to use full training data (edges/triplets) to generate robust node representations. The hypergraph datasets do not come with raw node features, and thus GNN baselines use randomly initialized features as input for training along with other model parameters. R-GCN uses \texttt{RGCNConv} layers that support message passing with multiple relation types between different types of nodes, where the edge types (relations) are used as input besides node features.

Subgraph-based models only use partial edges/triplets for training. For SEAL, 1-hop enclosing subgraphs are extracted online during the training and inference. Then, it applies three GCN layers of 32  hidden dimensions plus a sort pooling and several 1D convolution layers to generate a readout of the target subgraph for prediction. SUREL consists of a 2-layer MLP for query-level relative position encoding (RPE) and a 2-layer RNN to encode joined walks with attached RPEs. The hidden dimension of both networks is set to 64. The obtained readout of joined walks is aggregated and fed into a 2-layer MLP classifier to make predictions. GDGNN employs \texttt{GINLayer} as its backbone to obtain node embeddings. The horizontal geodesic representation is used for predictions, which finds the shortest path between two nodes in a query and aggregates node representations generated by GNNs along the found geodesic path. The max search distance for geodesic is the same as the number of GNN layers. For \texttt{collab}, \texttt{ppa}, \texttt{citation2} and \texttt{vessel}, the threshold of distance is set to 4, 4, 3, and 2, respectively. The hidden dimension of all fully connected layers is set to 32.

\section{Architecture and Hyperparameter}
\proj uses a 2-layer MLP with ReLU activation for encoding structural features and supports three set neural encoders, including mean pooling, LSTM, and attention. LSTM interprets elements to be aggregated in a set as a sequence \cite{hamilton2017inductive}; attention first calculates soft attention scores for elements in a set and then performs attention-score-weighted average pooling. The hidden dimension of all parameterized layers is set to 96. Lastly, hidden representations of query-level joined node sets are fed into a 2-layer MLP classifier for final predictions. 

The walk-based sampler builds on the sampling function from \texttt{SubGAcc}\footnote{\url{https://github.com/VeritasYin/subg_acc}} library developed by the authors, which also provides the support for efficient structural feature compression and index remapping. The metric-based sampler is adopted from fast PPR approximation in \cite{bojchevski2020scaling}.

\begin{table}
\centering
\caption{Hyperparameters Used for Benchmark \proj.\label{tab:hyper}}
\vspace{-2mm}
\resizebox{\columnwidth}{!}{
    \begin{tabular}{lccp{1.7cm}p{1.5cm}p{1.5cm}}
        \toprule
        \textbf{Dataset} & \#steps $m$ & \#walks $M$ & \#negative samples $k$ & Structural Feature & Set Neural Encoder\\
        \midrule
        \texttt{criteo-click} & 4 & 200 & 10 & LP & Mean\\
        \texttt{twitter-2010} & 4 & 100 & 25 & LP & Mean\\
        \texttt{citation2} & 4 & 100 & 10 & LP & Mean\\
        \texttt{collab} & 3 & 200 & 10 & LP & Mean\\
        \texttt{ppa}  & 4 & 200 & 20 & LP & Attn.\\
        \texttt{vessel} & 2 & 50 & 5 & LP & Mean\\
        \texttt{MAG (P-A)} & 3 & 200 & 10 & LP & Mean\\
        \texttt{MAG (P-P)} & 4 & 100 & 10 & LP & Mean\\
        \texttt{tags-math}  & 4 & 200 & 10 & LP & Mean \\
        \texttt{DBLP-coauthor} & 3 & 100 & 10 & LP & Mean \\
      \bottomrule
    \end{tabular}}
    \vspace{-2mm}
\end{table}

We follow the inductive setting for link and relation prediction: only partial samples will be used for training. Over the training graph, we randomly select 5\% links as positive training queries, each paired with $k$-many negative samples ($k=10$ by default). We mask these links and use the remaining 95\%  links to compute each node's structural features in the split training set via structure encoder. For \texttt{vessel}, as the input graph is very sparse, we first sort the nodes in training set by their degree and then randomly pick 5\% nodes to obtain edges of their 2-hop induced subgraphs for training and the rest reserved for structural feature construction. For higher-order pattern prediction, we use the given graph before timestamp $t$ to sample node sets and encode their structural features. The model parameters are optimized by triplets provided in the training set. No node features are used in \proj, except for \texttt{vessel} where normalized physical locations of each node are attached after its structural features and similarly for contextual features in \texttt{click}.

Table \ref{tab:cost_full} presents the extended version of Table \ref{tab:cost}. The results reported in Table \ref{tab:ogb} and the profiling of \proj in Tables \ref{tab:cost}, \ref{tab:cost_full} are obtained through the combination of hyperparameters listed in Table \ref{tab:hyper}. The dropout rate on \texttt{vessel} is set to \texttt{p=0.2}. The metric-based sampler is adopted to obtain the results of using PPR and SPD as structural features in Table \ref{tab:param}. Its sampling size $K$ is set to $50$, $50$ and $150$ for \texttt{collab}, \texttt{ppa}, \texttt{citation2}, respectively. The walk-based sampler is used for the results of LP as structural features, whose sampling parameters are listed in Table \ref{tab:hyper}. The rest of the hyperparameters remain the same as reported in Sec. \ref{sec:exp}. The \proj framework including \texttt{SubGAcc} library is open-source and free for academic use under the BSD-2-Clause license.

\end{document}